\documentclass{article} %
\usepackage[preprint]{neurips_2026}

\usepackage{algorithm}
\usepackage{algpseudocode}  %
\usepackage{wrapfig}
\usepackage{float}          %
\usepackage[utf8]{inputenc} %
\usepackage[T1]{fontenc}    %
\usepackage{hyperref}       %
\usepackage{url}            %
\usepackage{booktabs}       %
\usepackage{enumitem}

\usepackage{amsfonts}       %
\usepackage{nicefrac}       %
\usepackage{microtype}      %
\usepackage[table]{xcolor}         %
\usepackage{enumitem}
\usepackage{xspace}
\usepackage{lineno}

\usepackage{mathtools}
\usepackage{todonotes}
\usepackage{amssymb,fge}
\usepackage{thm-restate}
\usepackage{wrapfig}
\usepackage{bbm}
\usepackage{mathrsfs}
\usepackage{soul}
\usepackage{array}
\usepackage{multirow}
\usepackage{algorithm}
\usepackage[commentColor=black,beginLComment=/*~, endLComment=~*/]{algpseudocodex}
\usepackage{subcaption}
\usepackage[font=small]{caption}
\usepackage{pifont}%
\usepackage{tikz-3dplot}
\usepackage{pgfplots}
\pgfplotsset{compat=newest}
\usepackage{tikzscale}
\usepackage{comment} %
\PassOptionsToPackage{final,pagebackref=true}{hyperref}
\usepackage{hyperref} %
\usepackage{cleveref}
\usepackage{wasysym}
\usepackage{acronym}
\usepackage{xfrac}
\usepackage{multibib}
\newcites{Supp}{Supplement References}

\usepackage{amsthm}
\usepackage{xcolor}
\usepackage{amsmath,amsfonts,bm}
\makeatletter
\newtheorem*{rep@theorem}{\rep@title}
\newcommand{\newreptheorem}[2]{%
\newenvironment{rep#1}[1]{%
 \def\rep@title{#2 \ref{##1}}%
 \begin{rep@theorem}}%
 {\end{rep@theorem}}}
\makeatother

\newreptheorem{theorem}{Theorem}

\definecolor{myred}{RGB}{215,48,39}
\definecolor{mygreen}{RGB}{26,152,80}

\newcommand{\halfmark}{\textcolor{gray}{\checkmark\kern-1.1ex\raisebox{.7ex}{\rotatebox[origin=c]{125}{--}}}}
\usepackage[framemethod=TikZ]{mdframed}
\mdfdefinestyle{MyFrame}{%
    linecolor=black,
    outerlinewidth=.3pt,
    roundcorner=5pt,
    innertopmargin=1pt, %
    innerbottommargin=1pt, %
    innerrightmargin=1pt,
    innerleftmargin=1pt,
    backgroundcolor=black!0!white}

\mdfdefinestyle{MyFrame2}{%
    linecolor=white,
    outerlinewidth=1pt,
    roundcorner=1pt,
    innertopmargin=3pt,
    innerbottommargin=2pt,
    innerrightmargin=7pt,
    innerleftmargin=7pt,
    backgroundcolor=black!3!white}

\mdfdefinestyle{MyFrameEq}{%
    linecolor=white,
    outerlinewidth=0pt,
    roundcorner=0pt,
    innertopmargin=0pt,
    innerbottommargin=0pt,
    innerrightmargin=7pt,
    innerleftmargin=7pt,
    backgroundcolor=black!3!white}

\newcommand{\RNum}[1]{\uppercase\expandafter{\romannumeral #1\relax}}
\newcommand{\mb}[1]{\mathbf{#1}}

\newcommand{\R}{\mathcal{R}}

\newcommand{\vertiii}[1]{{\left\vert\kern-0.25ex\left\vert\kern-0.25ex\left\vert #1 
    \right\vert\kern-0.25ex\right\vert\kern-0.25ex\right\vert}}
\newcommand{\vertiiii}[1]{{\vert\kern-0.25ex\vert\kern-0.25ex\vert #1 
    \vert\kern-0.25ex\vert\kern-0.25ex\vert}}

\usepackage{mathtools}

\newcommand{\xhdr}[1]{{\noindent\bfseries #1}.}
\newcommand{\cut}[1]{}

\newcommand{\removelatexerror}{\let\@latex@error\@gobble}

\def\eqref#1{Eq.~\ref{#1}}

\def\1{\bm{1}}

\DeclareMathAlphabet{\mathsfit}{\encodingdefault}{\sfdefault}{m}{sl}
\SetMathAlphabet{\mathsfit}{bold}{\encodingdefault}{\sfdefault}{bx}{n}

\def\gL{{\mathcal{L}}}

\def\gN{{\mathcal{N}}}

\def\gP{{\mathcal{P}}}

\def\R{{\mathbb{R}}}

\newcommand{\sethree}{\mathrm{SE(3)}}

\newcommand{\namelong}{\textsc{\textbf{De}noiser \textbf{C}ofolding \textbf{A}ll-atom \textbf{F}lowmap}\xspace}
\newcommand{\nameshort}{\textsc{DeCAF}\xspace}
\newcommand{\namealg}{\textsc{DeCAF-SEARCH}\xspace}
\newcommand{\alphafold}[1]{AlphaFold~#1\xspace}
\newcommand{\af}[1]{AF#1\xspace}
\newcommand{\boltz}[1]{\mbox{Boltz-#1\xspace}}
\newcommand{\chai}[1]{\mbox{Chai-#1\xspace}}
\newcommand{\protenix}{\mbox{ProteniX\xspace}}
\newcommand{\pearl}{Pearl\xspace}
\newcommand{\pearldev}{Pearl-2026.1.dev\xspace}

\newcommand{\rmsdtwoa}{\text{RMSD\,\raisebox{0.25ex}{\scriptsize <}\,2\,\AA}\xspace}

\newcommand{\posebusters}{PoseBusters\xspace}
\newcommand{\runsnposes}{Runs~N'~Poses\xspace}
\newcommand{\rnp}{RnP\xspace}

\newcommand*{\backrefalt}[4]{%
    \ifcase #1 \footnotesize{(Not cited.)}%
    \or        \footnotesize{(Cited on page~#2)}%
    \else      \footnotesize{(Cited on pages~#2)}%
    \fi}
\newcolumntype{P}[1]{>{\centering\arraybackslash}p{#1}}

\hypersetup{
	colorlinks=true,       %
	linkcolor=blue,        %
	citecolor=blue,        %
	filecolor=magenta,     %
	urlcolor=blue         
}

\title{Few-step Cofolding with All-Atom Flow Maps}

\author{Gianluca Scarpellini $^{1}$\thanks{Correspondence to \texttt{gianscarpe@genesistherapeutics.ai, joey.bose@imperial.ac.uk}}, Ron Shprints$^2$, Peter Holderrieth$^2$, Juno Nam$^2$, \\
\textbf{Pranav Murugan$^1$}, \textbf{Rafael G\'omez-Bombarelli$^2$}, \textbf{Tommi Jaakkola$^2$}, \\
\textbf{Maruan Al-Shedivat$^1$},
\textbf{Nicholas Matthew Boffi$^3$}, \textbf{Avishek Joey Bose$^{1,4,5}$} \\
$^1$Genesis Molecular AI, $^2$ Massachusetts Institute of Technology, \\ 
$^3$ Carnegie Mellon University, $^4$ Imperial College London, $^5$Mila,
}

\begin{document}

\maketitle

\begin{abstract}

\looseness=-1

All-atom generative modeling of $3\mathrm{D}$ biomolecular complexes has emerged as the dominant paradigm for predicting the structure of proteins and protein-ligand systems.
Generating structures at the atomic level of fidelity, however, typically requires expensive iterative diffusion rollouts, making both conventional deployment and inference-time search techniques computationally costly.
In this paper, we introduce the \namelong (\nameshort) framework for distilling state-of-the-art all-atom cofolding models into all-atom flow maps that produce high-quality samples in only a few inference steps.
We build \nameshort on a denoiser-based formulation of flow maps with endpoint losses that naturally support $\sethree$ rigid alignment, which we show is critical for training accurate models.
We further derive a simple change of variables that lets \nameshort operate in the $\sigma$-space noise schedule of EDM-style architectures, enabling direct distillation from pretrained cofolding diffusion models.
Equipped with \nameshort's flowmap lookahead, we introduce a purpose-built inference-time framework that improves sampling through reward-guided search.
Empirically, \nameshort-Boltz statistically improves over \boltz{1x} in both accuracy (RMSD) and physical validity scores of protein-ligand poses at strict NFE budgets on the challenging \runsnposes, while also showing a more optimal Pareto frontier across all inference compute budgets on \posebusters. Distilling the state-of-the-art \pearl cofolding model, \nameshort-Pearl outperforms diffusion-based cofolding models and matches its teacher on success rate while using $5\times$ fewer NFEs.
We release our code at \url{https://github.com/genesistherapeutics/decaf}.

\end{abstract}

\section{Introduction}
\label{sec:introduction}

\looseness=-1
The accurate and efficient computational modeling of biological complexes has the potential to transform both our understanding of biomolecular mechanisms and our ability to catalyze the rational design of novel therapeutics~\citep{chevalier2017massively,ebrahimi2023engineering}.
At the core of this challenge is the need to model the intricate $3\mathrm{D}$ intermolecular structures that govern processes such as protein folding, protein--ligand binding, and biocatalysis.
This perspective is the thesis of structure-based drug design (SBDD), which seeks to engineer molecular structure to impart a desired downstream biological \emph{function}.
Despite its promise, progress in traditional SBDD is limited by the cost and latency of experimental structure determination.
In contrast, AI-driven design offers a promising alternative by instead leveraging scalable computational approaches to unlock discoveries of novel therapies~\citep{silva2019novo,murray2022novo,strauch2017computational,gainza2020deciphering,cao2021denovocovid}. 

\looseness=-1
The modern \emph{de facto} standard for AI-based SBDD is underpinned by large-scale generative models that represent biomolecular complexes of experimentally resolved structures directly at all-atom resolution, beginning with \alphafold{3}~\citep{abramson2024accurate} and followed by 
other luminary works~\citep{chai2024chai1,wohlwend2025boltz,boltz2,protenixv1,genesis2025pearl}.
This perspective is well-suited to learning the global geometry of the target distribution, reflected in structural features such as relative pose, backbone arrangement, and secondary structure.
However, unlike in classical generative modeling domains, success in biomolecular generation fundamentally requires modeling fine-grained local structure.
Indeed, in high-impact application settings like protein-ligand cofolding, inaccurate local structure modeling leads to catastrophic failure modes, yielding physically invalid generations that often include steric clashes, incorrect bond lengths and angles, strained side chain placements, and other stereochemical artifacts~\citep{wohlwend2025boltz}.

\looseness=-1
Strict physical and biological constraints have shifted much of the compute burden to \emph{inference time methods}.
In particular, generating structures requires expensive and fine-grained numerical simulation of the learned dynamics, which are needed to accurately predict local structure. In addition, for downstream testing, it remains essential to generate a diverse pool of candidates that increases the transfer rate from in-silico design to wet lab success. Furthermore, refining samples via inference-time search with proxy physical reward models is a critical aspect of the evolving protein generative pipeline and plays an important role in facilitating utility in high-impact downstream applications.
Despite this appeal, scaling inference is not a silver bullet and compounds the already expensive cost of numerical integration.
For instance, reward functions in the biological setting can often be expensive to query and are only applicable to fully denoised 3D structures.
Moreover, employing popular inference-time techniques that leverage multiple particles, such as Sequential Monte Carlo (SMC)~\citep{Del-Moral:2006}, Feynman-Kac steering (FK)~\citep{singhal2025general}, and Monte-Carlo Tree Search (MCTS)~\citep{jain2025diffusion}, doubly inflict the inference tax as they require multiple reward queries over each particle during the inference trajectory.
This raises the natural motivating research question: 

\begin{center}
\vspace{-5pt}
    \textit{\textbf{Q.} Can we train an all-atom cofolding model that can generate reward-optimized samples efficiently, using only a small number of neural function evaluations (NFEs) at inference time?}
\vspace{-5pt}
\end{center}

\begin{figure}[t]
  \centering
   \includegraphics[width=0.95\linewidth]{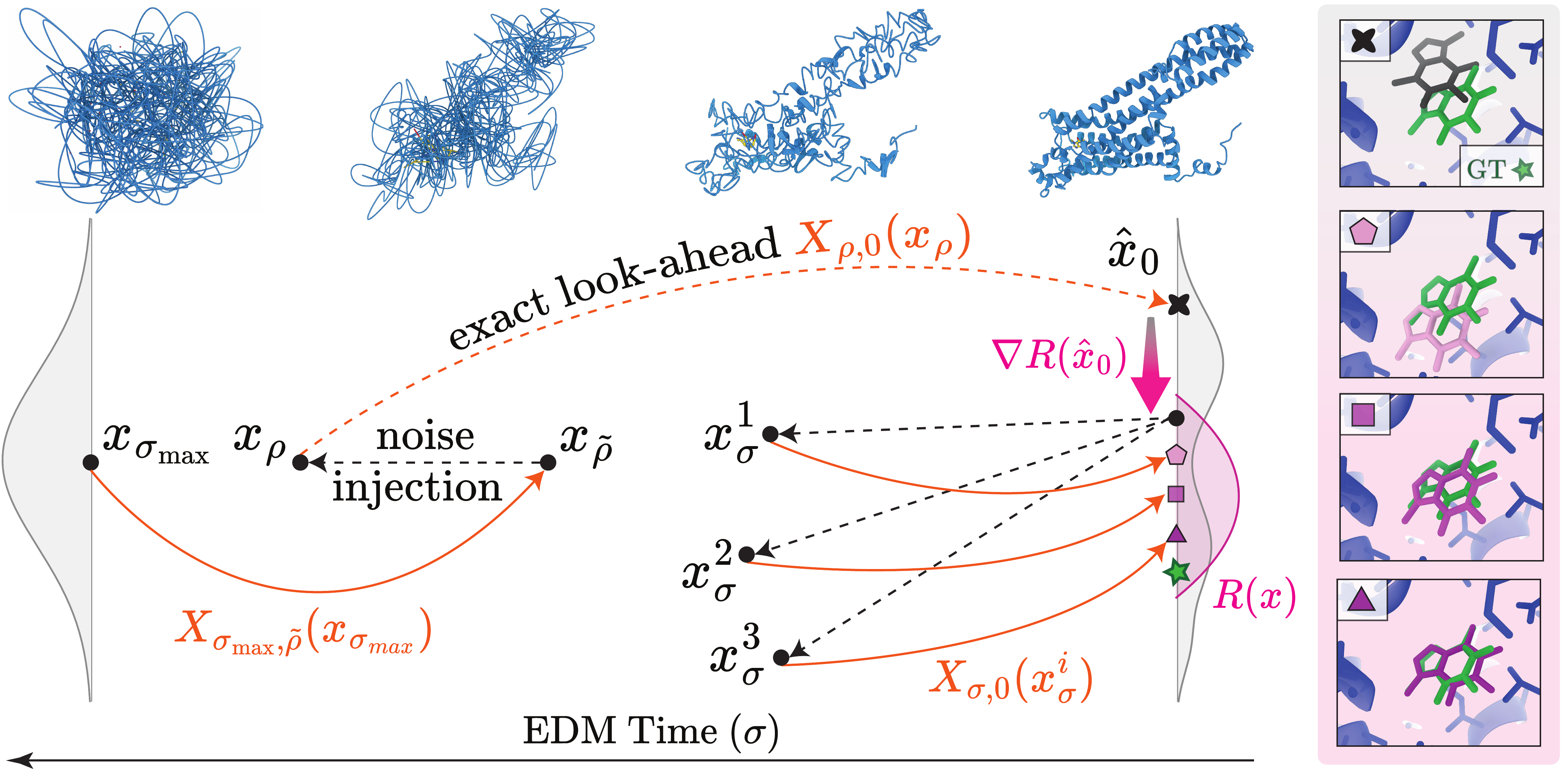}

  \caption{\small \looseness=-1 \nameshort accelerates all-atom biomolecular structure prediction with a few-step flow-map lookahead across EDM noise levels.
Atom-resolution guidance enables candidate search toward high-reward configurations.}
    \vspace{-0.5cm}
  \label{fig:visual_abstract}
\end{figure}

\looseness=-1
\xhdr{Main contributions}
In this paper, we answer the above question affirmatively and introduce the \namelong (\nameshort) framework. \nameshort is built on the flow map framework~\citep{flowmaps,boffi2024flow} (\Cref{fig:visual_abstract}) and efficiently distills a pretrained \boltz{1} model into the first all-atom cofolding model. Overall, we summarize our contributions as follows:

\begin{enumerate}[topsep=0pt, partopsep=0pt, itemsep=0pt, parsep=0pt, leftmargin=*]
\item \looseness=-1 \xhdr{Training methods} We design the first all-atom protein flow-map with \nameshort-Boltz, and outline key methodological innovations that enable effective distillation of a pre-trained \boltz{1} teacher (we refer to \nameshort for short when the pre-trained teacher is clear from the context).
In particular, we construct a novel reparameterization of the flow map in $\sigma$ space that allows an easy conversion of standard flow map objectives for all-atom modeling. 
\nameshort further exploits a denoiser-based flow map parametrization that crucially enables an $\sethree$ weighted rigid alignment of the ground truth structure---softly enforcing $\sethree$ symmetry---that is critical for stable training.
\item \looseness=-1 \xhdr{Inference methods} We introduce \namealg, an inference-time framework that leverages \nameshort's flow map lookahead that enables higher fidelity reward estimation in comparison to all-atom diffusion models. \namealg shares the benefits of stochastic sampling~\citep{kim2023consistency}, diffusion-SMC samplers~\citep{singhal2025general}, Diffusion MCTS~\citep{jain2025diffusion}, Diamond maps~\citep{holderrieth2025glass, potaptchik2026meta}, and FMRG~\citep{huang2026guide} while being fundamentally an \emph{inference-time search} method generating physically valid structures.
\item \xhdr{Empirical performance} We find that \nameshort-Boltz outperforms \boltz{1x} \emph{at low NFE budgets} on \runsnposes and matches the full-budget \boltz{1x} (600 NFE) on \posebusters with $20\times$ less compute at inference. Finally, we further apply \nameshort to a state-of-the-art cofolding model in \pearl~\citep{genesis2025pearl} and find \nameshort-\pearl improves over \boltz{1} and outperforms other potent baselines with full simulation while requiring $\approx 5-20\times$ fewer model evaluations.

\end{enumerate}

\looseness=-1

\section{Background, Preliminaries, and Related Work}
\label{sec:background}

\subsection{All-Atom Biomolecular Diffusion Modeling}
\label{sec:all_atom_background}

\looseness=-1
The dominant paradigm for all-atom biomolecular generative modeling is centered around diffusion-based cofolding models such as \alphafold{3}~\citep{abramson2024accurate}, which operates directly on the native Euclidean coordinates of structures. Through learning, \af{3} like models can perform the task of \emph{cofolding}---i.e., simultaneously predict the structure of a protein and a bound ligand. Standard implementations follow the EDM parameterization~\citep{karras2022elucidating}, which we review below.

\looseness=-1
\xhdr{VE Process}
Given a target distribution of all-atom structures $p_{\text{data}}(x) \in \gP(\R^d)$, the variance-exploding noising process corrupts a sample $x_0 \sim p_0(x_0):= p_{\text{data}}(x)$ with additive Gaussian noise,
\begin{equation}
    x_{\sigma} = x_0 + \sigma \epsilon,
    \qquad
    \epsilon \sim \gN(0,I),
    \qquad
    p_{\sigma}(x_\sigma) =p_0(x_0) * \gN(0,\sigma^2 I),
    \label{eq:ve_noising_kernel}
\end{equation}
\looseness=-1
where $\sigma \in [\sigma_{\min},\sigma_{\max}]$ indexes the noise level such that at times $t=0$ and $t=1$ there is $\sigma_{\min}$ and $\sigma_{\max}$ amounts of corruption added to $x_0$ respectively. Generation proceeds by solving for the time-reversal of the forward VE dynamics, which can be numerically simulated by following an ODE or SDE from high to low noise. For instance, we can simulate the probability-flow ODE: $  d x_{\sigma} / d \sigma
= - \sigma^2 \nabla_{x_{\sigma}} \log p_{\sigma}(x_{\sigma})$.
These reverse ODE dynamics require the Stein score $\nabla_{x_{\sigma}}\log p_{\sigma}(x_{\sigma})$, which can be estimated through a diffusion model's denoiser $D$ via Tweedie's formula~\citep{tweedie1957statistical}:
\begin{equation}
    s_{\sigma}(x_{\sigma})
    :=
    \nabla_{x_{\sigma}} \log p_{\sigma}(x_{\sigma})
    \approx
    \frac{D_{\sigma}(x_{\sigma})-x_{\sigma}}{\sigma^2}.
    \label{eq:score_denoiser_relation}
\end{equation}
\looseness=-1
\eqref{eq:score_denoiser_relation} can then be substituted in the flow ODE to simulate the reverse dynamics. The denoiser itself can be learned by a \emph{simulation-free} $\ell_2$-regression objective that performs denoising score matching across all noise levels ith a noise-dependent weighting function $\lambda(\sigma)$ against a clean sample $x_0$,
\begin{equation}
    \gL(\theta)
    =
    \mathbb{E}_{\sigma,\,x_0,\,\epsilon}
    \left[
        \lambda(\sigma)
        \left\|
            D_{\theta, \sigma}(x_0+\sigma\epsilon)-x_0
        \right\|_2^2
    \right],
    \qquad
    \epsilon \sim \gN(0,I).
    \label{eq:edm_denoising_loss}
\end{equation}
\looseness=-1
\xhdr{Structure prediction head}
For biomolecular all-atom diffusion models, the denoiser commonly leverages the EDM parametrization~\citep{karras2022elucidating} that designs a $\sigma$-dependent preconditioning:
\begin{equation}
    \hat{x}_{0}^{\mathrm{EDM}}
    :=
    D_{\theta, \sigma}(x_{\sigma})
    =
    c_{\mathrm{skip}}(\sigma)\,x_{\sigma}
    +
    c_{\mathrm{out}}(\sigma)\,
    F_{\theta, \sigma}\!\left(
        c_{\mathrm{in}}(\sigma)\,x_{\sigma},
        c_{\mathrm{noise}}(\sigma)
    \right).
    \label{eq:edm_parameterization}
\end{equation}
\looseness=-1
where $F_{\theta}$ is the raw network, $c_{\mathrm{skip}}$ controls the skip connection, $c_{\mathrm{in}}$ and $c_{\mathrm{out}}$ normalize input and output magnitudes, and $c_{\mathrm{noise}}$ embeds the noise level. In practice, the denoiser is implemented as a structure prediction head, typically a diffusion transformer with atom-attention encoder-decoder blocks.

\looseness=-1
The basic loss in~\eqref{eq:edm_denoising_loss} is augmented through $\sethree$ weighted rigid alignment of the ground truth structures to the prediction $\hat{x}^{\text{EDM}}_0$. This crucial step serves to simultaneously enforce soft global $\sethree$ symmetry and also reduce the variance of the diffusion loss. Finally, additional loss terms, including smooth-LDDT or bond-geometry penalties, encourage generating chemically plausible structures

\subsection{Flow Maps}
\looseness=-1
To accelerate inference in diffusion models, rather than simulating infinitesimal dynamics, one can learn a jump operator that directly traverses the probability-flow ODE associated with the diffusion model~\citep{song2023consistency,song2023improved, flowmaps,boffi2024flow}. This operator is known as the \emph{flow-map} and constitutes a map $X_{s,t}: [0,1]^2 \times \R^d \to \R^d$, which is the unique solution to the ODE that evolves the state dynamics between times $s$ and $t$---i.e., the jump condition satisfies $X_{s,t} (x_s) = x_t$, for all $(s,t) \in [0,1]^2$. This leads to a natural parametrization of the flow-map as a displacement using the \emph{average velocity} $u_{s,t}(x_s)$, between the two time points $s$ and $t$:
\begin{equation}
    X_{s,t}(x_s) = x_s + (t-s) u_{s,t}(x_s), \quad u_{s,t} (x_s) = \frac{1}{t-s}\int^t_s v_\tau(x_\tau) d\tau,
    \label{eq:flow_map parametrization}
\end{equation}
\looseness=-1
where $v_{\tau}(x_\tau)$, is the instantaneous velocity. It is clear from~\eqref{eq:flow_map parametrization}, that in the limit where the two time points converge  the average velocity recovers the instantaneous velocity of the ODE, $\underset{s\to t}{\lim} \partial_t X_{s,t}(x_s) = u_{t,t}(x_t) := v_t(x_t)$. This is known as the tangent condition~\citep{flowmaps}, and demonstrates that the flow-map contains an implicit instantaneous velocity in its parametrization. 

\looseness=-1
At optimality, the flow-map simultaneously enforces the following consistency rules:
\begin{equation*}
    \underbrace{\partial_t X_{s,t}(x_s) = u_{t,t}(X_{s,t}(x_s))}_{\text{Lagrange}}, \ \underbrace{\partial_s X_{s,t}(x_s) =- u_{s,s}(x_s) \nabla X_{s,t}(x_s)}_{\text{Euler}}, \ \underbrace{X_{r, t}(X_{s,r}(x_s)) = X_{s,t}(x_s)}_{\text{Progressive}}.
\end{equation*}
\looseness=-1
Each consistency condition naturally gives rise to a PINN-style loss function that facilitates learning the flow-map~\citep{flowmaps,boffi2024flow,shortcut, consistencyTrajectory}. Furthermore, these losses can be employed to either self-distill or distill a pre-trained diffusion model into a flow-map by computing the RHS of each consistency condition with a frozen pre-trained model.

\section{Method}
\label{sec:method}

\looseness=-1
 We now introduce \namelong(\nameshort), a novel $\sigma$-space flow-map framework that constructs an all-atom protein flow map by distilling a pre-trained all-atom teacher in the vein of \af{3}~\citep{abramson2024accurate}. In particular, \nameshort outputs a few-step generative model that captures the $3\mathrm{D}$ structure of protein-ligand interactions for the task of cofolding. Critically, \nameshort offers two principal advantages over its pre-trained all-atom teacher: 
 \begin{enumerate}[topsep=0pt, partopsep=0pt, itemsep=0pt, parsep=0pt, leftmargin=*]
\item \looseness=-1 \emph{\underline{Few-step inference:}} \nameshort offers accelerated inference that compresses the full simulation of a diffusion trajectory into a few denoising steps without compromising sample quality.
\item \looseness=-1 \emph{\underline{Flowmap lookahead:}} \nameshort by construction defines a lookahead map over end points that allows higher fidelity terminal reward estimation than a denoiser of an EDM model. As a result, any inference-time technique for reward optimization benefits not only fewer simulation steps that reduce simulation latency but also improved reward estimation and alignment.
 \end{enumerate}

\looseness=-1
We organize the remainder of the section as follows: in~\S\ref{sec:denoiser_mf}, we introduce our primary learning framework that exploits a denoiser-parametrization to learn \nameshort. In~\S\ref{sec:inference_time_steering}, we exploit our \nameshort to unlock new, more efficient mechanisms for inference-time search and reward alignment.

\subsection{\namelong}
\label{sec:denoiser_mf}

\looseness=-1
We first highlight several key technical challenges prevalent in the EDM parametrization that prevent the distillation of all-atom teacher models. EDM's residual form (\eqref{eq:edm_parameterization}) is conditioned on a single $\sigma$ and does not extend cleanly to dual-time flow maps. Importantly, this has already motivated the simplification of the EDM parametrization when distilling to flow-maps~\citep{fastgen2026}. In addition, for practical noise schedules, the monotone time reparameterization $\sigma: [0,1] \to [\sigma_{\mathrm{min}}, \sigma_\mathrm{max}]$ is non-linear, making loss computations that require time partials numerically unstable~\citep{ayf}. More precisely, when computing the chain rule $\partial x / \partial t = \left(\partial x / \partial \sigma \right) \left( \partial \sigma/ \partial t \right)$ for the suggested EDM noise schedule $\sigma_t$~\citep{karras2022elucidating}, the end points are prone to large magnitudes of $| \partial \sigma/ \partial t |$. This, in turn, leads to numerical instability when computing flow-map-based objectives. 

\looseness=-1
To circumvent the numerical instability of flow-map training using a pre-trained EDM protein model, we directly define a flow-map in the $\sigma$-noise space, allowing us to redefine all objectives and sampling directly over $\sigma$-steps---thus eliminating the problematic factor $ \partial \sigma/ \partial t$. This leads to our notion of $\sigma$-velocity, which represents the $\sigma$-instantaneous velocity along the PF-ODE,
\begin{equation}
      v_{\sigma}(x_{\sigma}) \;\triangleq\; \frac{d x_{\sigma}}{d\sigma}, \quad  v_{\sigma}^{\text{EDM}}(x_{\sigma}) = \frac{x_{\sigma} - D_{\sigma}(x_{\sigma})}{\sigma}= \frac{x_{\sigma} - \hat{x}^{\text{EDM}}_{0}}{\sigma}.
    \label{eq:teacher_sigma_vel}
\end{equation}
\looseness=-1
Our reparametrization of time extends, in a natural way, to now a two-noise level map that denoises $\rho \to \sigma$, for $\rho > \sigma$, using the analogous notion of \emph{average} $\sigma$-velocity and flow-map parametrization $X_{{\rho}, {\sigma}}(x_{\rho})$. This forms the basis of the sampling update at inference:
\begin{equation}
      u_{{\rho},{\sigma}}(x_{\rho}) = \frac{1}{\rho-\sigma}\int_{\rho}^{\sigma} v_{\bar{\sigma}}(x_{\bar{\sigma}})d\bar{\sigma}, \quad X_{{\rho}, {\sigma}}(x_{\rho}) = x_{\rho} - (\rho - \sigma)u_{{\rho},{\sigma}}(x_{\rho}).
      \label{eq:sigma_average_velocity}
\end{equation}
\looseness=-1
\xhdr{Denoiser Parametrization}
To train our all-atom protein flow map, we follow the mean-flow objective~\citep{meanflows}, which is also equivalent to the Eulerian objective~\citep{boffi2024flow}. This requires the construction of the instantaneous $\sigma$-velocity \emph{implied} by the average $\sigma$-velocity of~\eqref{eq:sigma_average_velocity}. Explicitly, we efficiently compute this quantity using Jacobian-vector products $\texttt{jvp}$,
\begin{align}
    v_{\rho}(x_{\rho}) &= u_{{\rho},{\sigma}}(x_{\rho}) + (\rho - \sigma) \frac{\mb{d}}{\mb{d} \rho} u_{{\rho},{\sigma}}(x_{\rho}), \label{eq:meanflow_total_derivative} \\    V(x_{\rho}, \rho, \sigma) & \triangleq u_{{\rho},{\sigma}}(x_{\rho}) + (\rho - \sigma) \cdot \texttt{sg}\left(\texttt{jvp}\left(u_{{\rho},{\sigma}}(x_{\rho}) , (x_{\rho}, \rho, \sigma), (v, 1, 0)\right)\right). \label{eq:meanflow_v_pred_jvp}
\end{align}
\looseness=-1
In~\eqref{eq:meanflow_total_derivative}, $\mb{d}/\mb{d}\rho$ represents the total derivative of the average $\sigma$-velocity, while $\texttt{sg}$ is the stopgrad operator applied to $\mb{d}u_{\rho, \sigma}/\mb{d}\rho$. 
In the case of distilling from a pre-trained EDM model, we simply substitute the ground-truth $\sigma$-velocity with $v^{\text{EDM}}_{\sigma}$. This leads to a natural learning objective of computing the predicted instantaneous $\sigma$-velocity using~\eqref{eq:meanflow_v_pred_jvp} and matching it to the EDM teacher velocity in~\eqref{eq:teacher_sigma_vel}~\citep{meanflows,geng2025improved,lu2026one,potaptchik2026discrete}. We, however, motivate a different approach in the context of all-atom protein models. Specifically, we highlight a standard practice in \af{3} diffusion models that leads to a lower variance loss estimate, which is to instead predict the target end-point $\hat{x}_\text{tgt}$ and find the closest rigid transformation---i.e., $\sethree$ rigid alignment---to the target $\hat{x}^{\text{EDM}}_0$. As a result, we parametrize our all-atom flow map as a denoiser that consumes two noise levels $D(x_{\rho}, \rho, \sigma)$. Specifically, we can recover a  $\hat{x}_\text{tgt}$-prediction by computing using the predicted $\sigma$-instantaneous velocity of~\eqref{eq:meanflow_v_pred_jvp}:  
\begin{equation}
  \hat{x}^{\text{\nameshort}}_\text{tgt} := D(x_{\rho}, \rho, \sigma) = x_{\rho} - \rho \cdot V(x_{\rho}, \rho, \sigma)
  \label{eq:denoiser_parameterization_flow_map}
  \end{equation}
\looseness=-1
The above equation gives rise to a \emph{two-time} denoiser~\citep{lu2026one,lee2026flow,potaptchik2026discrete,roos2026categorical} that can be leveraged to construct an endpoint loss:
\begin{mdframed}[style=MyFrameEq]
\begin{equation}
\mathcal{L}
=
\mathbb{E}_{x_{0}, x_{\rho}, x_{\sigma}} \!\left[\frac{1}{\sigma^2} \!\left[
\min_{g \in \sethree}\left\|
 \hat{x}^{\text{\nameshort}}_{\text{tgt}} - \texttt{sg}\left(\zeta(g) \circ \hat{x}^{\text{EDM}}_{0}\right)
\right\|^2
\right]\right].
\label{eq:main_loss}
\end{equation}
\end{mdframed}
\looseness=-1

Here we take argmin over the entire group $\sethree$: $\zeta(g)$ is its matrix representation and is the rigid alignment step performed using the Kabsch algorithm, while $\hat{x}^{\text{EDM}}_{0}$ is the prediction computed by the pre-trained EDM teacher. 
We emphasize that this exact formulation fails under a velocity loss, as subtracting translation would lose a degree of freedom, and thus complicates the distillation setup. 

\looseness=-1

The loss in~\eqref{eq:main_loss} naturally supports both off-diagonal training and diagonal training through sampling of noise levels $(\rho, \sigma) \sim \text{Sampler}(\sigma_{\text{min}}, \sigma_{\text{max}})$. In particular, when $\rho = \sigma$, we learn to match exactly the score associated with the PF-ODE of the EDM teacher, while in all other off-diagonal cases, we learn to take $(\rho - \sigma)$ jumps along the trajectory of the PF-ODE as in~\eqref{eq:sigma_average_velocity}.

\looseness=-1
\xhdr{Flow map sampling algorithm}
To sample from \nameshort we design a stochastic $\gamma$-sampler, which shares inspiration from the consistency models literature~\citep{kim2023consistency}. We present this $\gamma$-sampler in~\cref{alg:gamma_sampler}, which allows us to toggle deterministic sampling ($\gamma=0$) to more stochastic sampling for $\gamma > 0$. As we later demonstrate in our experiments~\S\ref{sec:experiments_main}, the added stochasticity aids overall performance at no added cost and also enjoys seamless integration with our inference strategy in~\S\ref{sec:inference_time_steering}.

\subsection{Inference-Time Search}
\label{sec:inference_time_steering}
\looseness=-1
\begin{wrapfigure}{r}{0.52\textwidth}
    \vspace{-40pt}
    \begin{minipage}{\linewidth}
      \begin{algorithm}[H]                  %
      \caption{\nameshort{} $\gamma$-sampling}
      \label{alg:gamma_sampler}
      \begin{algorithmic}[1]
        \Require FlowMap $X$; $N$ steps; $\gamma\in[0,1]$.
        \State $x_{\sigma_N} \sim \mathcal{N}(0,\,\sigma_N^2 I)$
        \For{$n = N$ \textbf{down to} $1$}
            \State $\tilde{\sigma}_{n-1} \gets \sqrt{1-\gamma^2}\,\sigma_{n-1}$
            \State $x_{\tilde{\sigma}_{n-1}} \gets
  X(x_{\sigma_n},\,\sigma_n,\,\tilde{\sigma}_{n-1})$
            \Comment{flowmap}
            \State $\epsilon \sim \mathcal{N}(0, I)$
            \State $x_{\sigma_{n-1}} \gets x_{\tilde{\sigma}_{n-1}} + \gamma\,\sigma_{n-1}\,\epsilon$
            \Comment{re-noise}
        \EndFor
        \State \Return $x_{\sigma_0}$
      \end{algorithmic}
      \end{algorithm}
    \end{minipage}
    \vspace{-10pt}
  \end{wrapfigure}
  
\looseness=-1
The challenge of generating physically plausible biomolecular structures has motivated a vast literature on computationally intensive inference-time correction of pre-trained all-atom diffusion models.
While \nameshort is primarily designed as a few-step all-atom model that reduces inference latency for conventional deployment, it also enables a second computational advantage. Specifically, \nameshort also enables substantial gains in \emph{inference-time search} (i.e., reward alignment~\citep{uehara2025inference}) for the cofolding problems we consider. 

\looseness=-1
In biomolecular modeling, a common form of inference-time search defines a terminal reward $R:\mathbb{R}^d\to\mathbb{R}$ that measures physical validity, for example, penalizing steric clashes or violations of bond lengths and angles~\citep[Section~4]{wohlwend2025boltz}.
The goal is to steer generation toward samples with high $R(x_0)$ while preserving structural accuracy. A central difficulty is that $R$ is only defined on clean structures, whereas steering decisions must be made from noisy intermediate states $x_\sigma$. Consequently, we require an efficient estimate of the reward expected after denoising $x_\sigma$ to a clean structure.
\nameshort provides such an estimate through its learned two-time flow map: given an intermediate state $x_\sigma$, we compute a look-ahead (end point) prediction over clean samples
\begin{equation}
    \hat{x}_0 = X(x_{\sigma}, \sigma, 0) = x_{\sigma} - \sigma \cdot u_{\sigma, 0}(x_{\sigma}),
    \label{eq:sample_parametrization}
\end{equation}
\looseness=-1
and use $R(\hat{x}_0)$ as a proxy for the expected reward of $x_\sigma$.  
Similar flow map look-aheads have been used for steering in the image domain~\citep{sabour2025test,holderrieth2026diamond,potaptchik2026meta,huang2026guide}. We explore their analogue in all-atom biomolecular generation, which differs in two important ways.
First, our base sampler is not a standard ODE or SDE integrator but $\gamma$-sampler.
Second, rewards are based on physical violations whose supervisory signal is highly non-smooth and uninformative at noisy states $x_\sigma$.
These considerations motivate refining generations through clean-space look-aheads rather than through gradients in noisy state space.

\looseness=-1
\xhdr{\namealg}
We introduce an inference-time search algorithm \namealg for all-atom flow maps (see~\cref{alg:fm_search}).
\namealg adapts standard inference-time steering methods, including Feynman--Kac (FK)~\citep{singhal2025general,skreta2025feynman} and diffusion-based MCTS~\citep{jain2025diffusion}, to flow maps and the $\gamma$-sampling setting. Starting from a population of particles, we repeatedly denoise each particle from $x_\sigma$ to a clean look-ahead $\hat{x}_0$ (possibly over several flow map steps), evaluate its reward, and optionally improve it by clean-space gradient ascent 
\begin{equation}
    \hat{x}_0 \leftarrow \hat{x}_0 + \beta \nabla_{\hat{x}_0} R(\hat{x}_0).
    \label{eq:gradient_steps}
\end{equation}
\looseness=-1
Then, we renoise the improved structure to a noisy state $x_{\tilde{\sigma}}$ before continuing to the next $\gamma$-sampling iteration. Inspired by \citet[Proposition 5.1.]{holderrieth2026diamond}, we also consider a variant where the gradient is an average of several Monte Carlo samples (MC-GRAD in~\S\ref{app:sampling_algos}). Across particles, compute is preferentially allocated to promising branches using either SMC resampling or an upper-confidence-bound criterion.
With finite-temperature resampling, \namealg resembles FK steering with UCB/UCT selection and recovers a simple MCTS-style variant~\citep{jain2025diffusion}.
\namealg also extends the FK-style steering used in \boltz{1x}~\citep{wohlwend2025boltz}: intermediate clean predictions are obtained using the learned \nameshort flow map, which can provide more accurate or more efficient look-aheads than a single-step denoiser.

\section{Experiments}
\label{sec:experiments_main}

\looseness=-1
We investigate the application of \nameshort for the task of cofolding protein-ligand interactions. In particular, we distill a pre-trained \boltz{1} model using~\eqref{eq:main_loss} and sample using \namealg (\cref{alg:fm_search}), in contrast to \boltz{1x} the physical potential variant~\citep{wohlwend2025boltz}.
For fair comparison, \nameshort shares architecture, training data, and pretraining with \boltz{1} (see~\S\ref{app:experimental_setup} for architecture, experimental setup, and hyperparameters). We include additional ablations in~\S\ref{app:additional_experiments}. 

\looseness=-1
\xhdr{Benchmarks}
The primary benchmark we use for analysis is  \runsnposes (RnP)~\citep{vskrinjar2025have}. We limit the experiments to a stricter set of 702 structures with a cutoff date of \texttt{2023-06-01}.
We also conduct additional analysis on \posebusters~\citep{posebusters} with cutoff date of \texttt{2021-10-01}, yielding $282$ structures that can be handled by \boltz{1} on a single A100-80GB GPU.

\looseness=-1
\xhdr{Metrics}
Following the standard practice, we report the following metrics:
(i) \emph{\underline{\rmsdtwoa}} defined as the percentage of the test structures for which root mean square distance between ground-truth and generated ligand poses is under 2\,\AA;
(ii) \emph{\underline{PB-Valid}} defined as the percentage of test structures that are physically valid according to the PoseBusters library;
(iii) \emph{\underline{lDDT-PLI}} defined as local distance difference test \citep{mariani2013lddt} on the short-range protein--ligand contacts within a 6\,\AA\ protein-ligand pocket, where side-chain atoms typically outnumber backbone atoms.
We also define \emph{\underline{Success Rate (\%)}} as a percentage of the test structures that satisfy \rmsdtwoa and PB-Valid criteria.

\begin{figure}[t]
    \centering
    \begin{minipage}[t]{0.59\textwidth}
      \captionof{table}{%
        \small \textbf{\nameshort{} vs \boltz{1x} on \runsnposes{} benchmark.}
        The best recipe at or below the NFE budget is reported per method averaged over $5$ poses.
        \textbf{Bold} marks the winner between \boltz{1x} and \nameshort{};
        stars indicate paired Wilcoxon signed-rank significance vs the
        other model (two-sided): $^{*}p{<}0.05$, $^{**}p{<}0.01$,
        $^{***}p{<}0.001$. $^\dagger$ Indicates numbers taken from~\citep{genesis2025pearl}}
      \label{tab:rnp}
      \small
      \setlength{\tabcolsep}{4pt}
       \resizebox{1\linewidth}{!}{
      \begin{tabular}{lrrr}
        \toprule
        \textbf{Method}
          & \textbf{PB Valid}$\uparrow$
          & \textbf{Success Rate}$\uparrow$
          & \textbf{lDDT-PLI}$\uparrow$ \\
        \midrule
        \rowcolor{gray!10}\multicolumn{4}{l}{\textit{$\leq$ 10 NFEs (2 steps)}} \\
        \quad \boltz{1x} (default)  &  0.0  &  0.0  &  0.0  \\
        \quad \boltz{1x} (tuned)    &  21.8  &  17.2  &  45.8  \\
        \quad \namealg              &  $^{*}\mathbf{24.0}$  &  $^{**}\mathbf{19.6}$  &  \textbf{46.4}  \\
        \midrule
        \rowcolor{gray!10}\multicolumn{4}{l}{\textit{$\leq$ 20 NFEs (5 steps)}} \\
        \quad \boltz{1x} (tuned)    &  75.9  &  50.3  &  59.8  \\
        \quad \namealg              &  $^{***}\mathbf{83.5}$  &  $^{***}\mathbf{57.4}$  &  $^{***}\mathbf{66.5}$  \\
        \midrule
        \rowcolor{gray!10}\multicolumn{4}{l}{\textit{$\leq$ 25 NFEs (6 steps)}} \\
        \quad \boltz{1x} (tuned)    &  73.4  &  51.9  &  61.8  \\
        \quad \namealg              &  $^{***}\mathbf{80.1}$  &  $^{***}\mathbf{57.0}$  &  $^{***}\mathbf{67.8}$  \\
        \midrule
        \rowcolor{gray!10}\multicolumn{4}{l}{\textit{$\leq$ 40 NFEs (10 steps)}} \\
        \quad \boltz{1x} (tuned)    &  86.6  &  55.4  &  64.2  \\
        \quad \namealg              &  $^{***}\mathbf{91.6}$  &  $^{***}\mathbf{61.8}$  &  $^{***}\mathbf{67.2}$  \\
        \midrule
        \rowcolor{gray!10}\multicolumn{4}{l}{\textit{$\leq$ 50 NFEs (12 steps)}} \\
        \quad \boltz{1x} (tuned)    &  86.3  &  56.5  &  63.7  \\
        \quad \namealg              &  $^{***}\mathbf{92.7}$  &  $^{***}\mathbf{62.6}$  &  $^{**}\mathbf{67.1}$  \\
        \midrule
        \rowcolor{gray!10}\multicolumn{4}{l}{\textit{$\leq$ 160 NFEs}} \\
        \quad \boltz{1x} (default, 40 steps)  &  88.5 & 58.7 & 67.6 \\
        \quad \namealg{} (MC-GRAD, 15 steps)    & $^{***}\mathbf{ 92.1 }$&  $^{***}\mathbf{63.7}$  & $^{***}\mathbf{ 69.2 }$ \\
                \midrule
        \rowcolor{gray!10}\multicolumn{4}{l}{\textit{800 NFEs (Full budget)}} \\
        \quad \boltz{1x} (ref., 200 steps)  &  \textbf{95.9}  &  64.9  &  \textbf{69.9}  \\
        \quad \namealg{} (MCTS, 10 steps)    &  95.5  &  \textbf{65.0}  &  69.5  \\
        \midrule
        \multicolumn{4}{l}{\textcolor{black!50}{\textit{Frontier cofolding models}$^\dagger$ }} \\
        \quad\textcolor{black!50}{\alphafold{3}}  &  \textcolor{black!50}{73.9}  &  \textcolor{black!50}{60.7}  &  \textcolor{black!50}{80.9}  \\
        \quad\textcolor{black!50}{\pearl}         &  \textcolor{black!50}{96.1}  &  \textcolor{black!50}{72.1}  &  \textcolor{black!50}{81.9}  \\
        \bottomrule
      \end{tabular}
      }
    \end{minipage}
    \hfill
    \begin{minipage}[t]{0.40\textwidth}
      \vspace{-5pt}
      \centering
      \includegraphics[width=.9\linewidth]{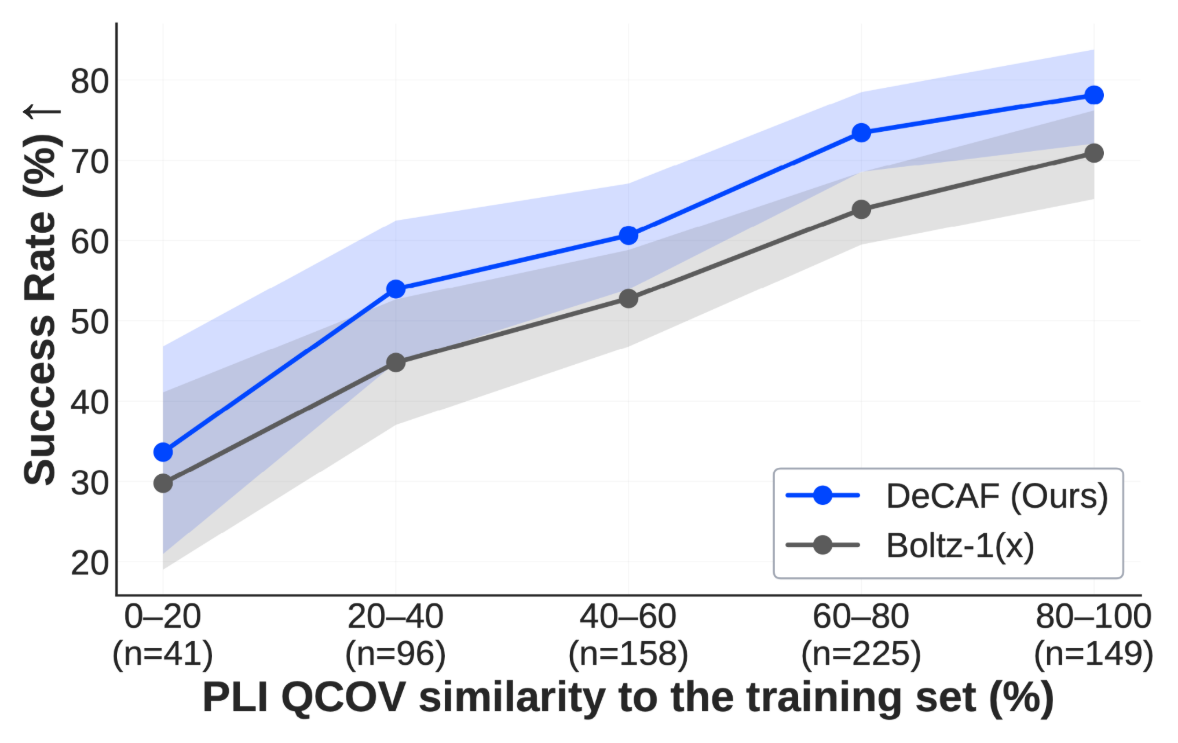}
      \vspace{-5pt}
      \captionof{figure}{\small Success rate vs.\ training-set similarity on the \rnp benchmark at 40 NFE. }
      \label{fig:rnp-pliqcov}
      \includegraphics[width=.9\linewidth]{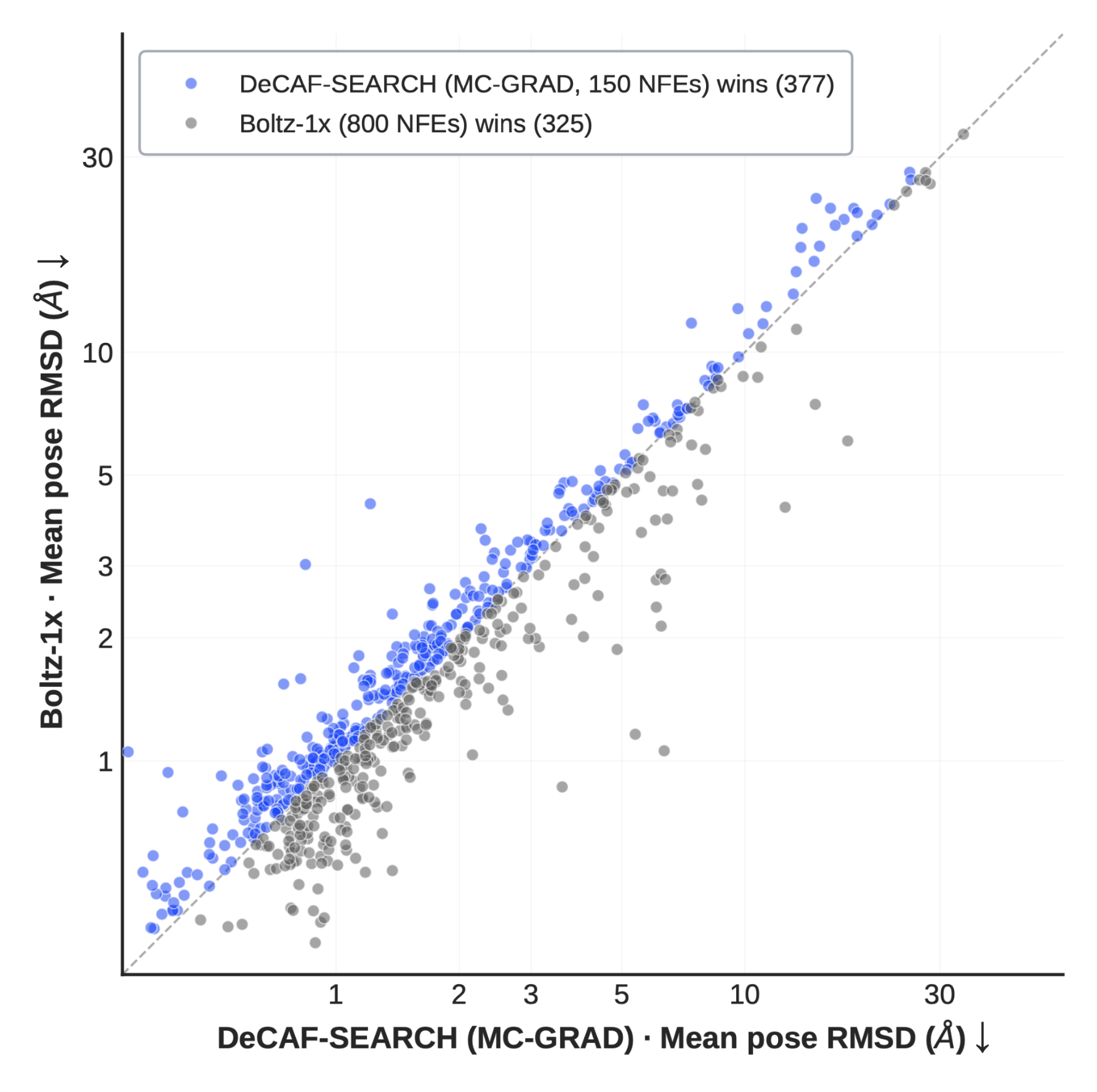}
      \vspace{-10pt}
      \captionof{figure}{\small \looseness=-1 Mean RMSD per structure for \namealg (150 NFEs) ($x$) vs. \boltz{1x} (800 NFEs) ($y$) on \rnp. }
      \label{fig:rnp-scatter}
    \end{minipage}
    \vspace{-18pt}
  \end{figure}

\looseness=-1
Our experiments seek to answer three key questions that test the empirical caliber of \nameshort:
\begin{itemize}[topsep=0pt, partopsep=0pt, itemsep=0pt, parsep=2pt, leftmargin=25pt]
      \item[\textbf{(Q1.)}] \looseness=-1 \textbf{Low NFE regime.} Does \nameshort{} outperform \boltz{1} with a limited inference budget (\S\ref{sec:exp:few_step})?
      \item[\textbf{(Q2.)}] \looseness=-1 \textbf{Analysis of compute-optimal frontier.} What is the Pareto frontier of \namealg against \boltz{1x} for inference time scaling across \emph{any} inference compute budget (\S\ref{sec:exp:compute_optimal})?
      \item[\textbf{(Q3.)}] \looseness=-1 \textbf{Performance analysis.} What are the relative quantitative and qualitative differences in generation quality at the sample-level of \nameshort in comparison to its teacher \boltz{1} (\S\ref{sec:exp:qual})?
  \end{itemize}
\looseness=-1

\subsection{Performance at low NFEs (Q1.)}
\label{sec:exp:few_step}

\looseness=-1
We evaluate \nameshort and \boltz{1} on \runsnposes for various NFE inference regimes. We report our main results in~\cref{tab:rnp} (average over 5 poses) on \runsnposes{} structures drawn entirely from PDB depositions released \emph{after} the \texttt{2023-06} cutoff date for training. Specifically, we sample \namealg with SMC resampling over $4$ particles---bearing similarity to FK steering~\citep{singhal2025general}---at compute budgets of $10$, $20$, $25$, $40$, and $50$ NFEs. We also include a high-budget ($800$ NFEs) comparison between \boltz{1x} and \namealg with a variation that is closest to MCTS. As~\cref{tab:rnp} demonstrates, \boltz{1x} fails catastrophically at generating plausible structures at low steps regime under the default sampling configuration. As an orthogonal contribution, we remedy this by tuning the step scale in the default \boltz{1x} to $\eta=1$ when $\leq 15$ diffusion steps, which restores stable SDE sampling in the few-step regime. We refer to this setting as \boltz{1x-tuned}.

\looseness=-1
At \emph{every} low compute budget in \cref{tab:rnp}, \nameshort{} outperforms \boltz{1x-tuned} on \emph{every} metric. The improvement is significant in the 20--160 NFE range (paired Wilcoxon signed-rank, $p<0.001$ on nearly all metrics). At 50 NFEs, \nameshort{} is also competitive with frontier cofolding models that use $\geq800$ NFEs---surpassing even \alphafold{3} on the Success Rate and PB-Valid metrics. At a matched budget of $800$ NFEs, \nameshort{} is on par with \boltz{1x}  on all three metrics.

\looseness=-1
\xhdr{Generalization}
An important criterion for cofolding models is their ability to handle difficult targets, as it is a proxy for model generalization beyond the training set. In~\cref{fig:rnp-pliqcov}, we plot the success rate against the pocket-similarity metric \emph{PLI Q-Coverage} (shaded regions show 95\% bootstrap confidence intervals (1000 resamples per bin). We observe that \nameshort{}'s margin over \boltz{1x} holds across every quartile, including the most out-of-distribution bin (lowest PLI Q-Coverage), confirming that the performance gain is not driven by structures that are close to the training set. 

\looseness=-1
Lastly, in~\cref{fig:rnp-scatter} we perform a fine-grained analysis of performance at the sample-level. In particular, we conduct a comparison between \namealg (MC-GRAD) at 150 NFEs and \boltz{1x} (800 NFEs) for each of the $702$ structures in \runsnposes. Each point gives one target's mean pose RMSD under \nameshort{} ($x$) and \boltz{1x} ($y$); points above the $y\!=\!x$ diagonal indicate \nameshort{} wins.
\nameshort{} matches \boltz{1x} on the per-target distribution of mean RMSD at $5.3\!\times\!$ less inference compute. The advantage holds in the difficult tail: among the targets where at least one method exceeds $10$,\AA, \nameshort{} attains the lower RMSD on $71\%$ of them.

\begin{minipage}[t]{\linewidth}
  \centering
  \captionof{table}{%
    \small \textbf{\texttt{Post-2023} \runsnposes{} comparison ($n{=}196$)}. AlphaFold3, Chai-1, and Boltz-1 were trained with \texttt{2021-09} cutoff date. Boltz-2 and \pearl models are trained on \texttt{2023-06} and \texttt{2023-12} cutoff dates, respectively. Bolding is reserved for only the $40$ NFE comparisons and not the frontier models at full budget.}
  \label{tab:rnp_pearl}
  \resizebox{\linewidth}{!}{%
  \begin{tabular}{l ccc ccc}
    \toprule
    & \multicolumn{3}{c}{\textbf{best@1}} & \multicolumn{3}{c}{\textbf{best@5}} \\
    \cmidrule(lr){2-4} \cmidrule(lr){5-7}
    Method & PB-valid & Success Rate & lDDT-PLI & PB-valid & Success Rate & lDDT-PLI \\
    \midrule
    \rowcolor{gray!10}\multicolumn{7}{l}{\textit{Full budget}} \\
    AF3 $^\dagger$  & 66.3 & 53.5 & 76.1 & 86.2 & 69.9 & 79.3 \\
    Chai-1$^\dagger$  & 71.2 & 49.5 & 69.9 & 90.3 & 63.3 & 72.9 \\
    Boltz-1x$^\dagger$  & 93.7 & 62.2 & 71.3 & 95.9 & 66.3 & 73.2 \\
    Boltz-2$^\dagger$ \texttt{(2023-06 cutoff)} & 67.9 & 50.1 & 74.3 & 79.1 & 58.2 & 77.3 \\
    \pearldev \texttt{(2023-12 cutoff)} & 96.7 & 67.8 & 73.0 & 98.5 & 77.0 & 78.9 \\
    \midrule
    \rowcolor{gray!10}\multicolumn{7}{l}{\textit{$40$ NFEs ($10$ steps)}} \\
    \pearldev \texttt{(2023-12 cutoff)} & 83.9 & 52.0 & 60.6 & 97.4 & 75.0 & \textbf{78.7} \\
    \nameshort-Boltz & 89.5 & 55.5 & 62.3 & 98.0 & 66.8 & 67.7 \\
    \textbf{\nameshort-\pearl} & \textbf{95.1} & \textbf{64.8} & \textbf{71.1} & \textbf{99.5} & \textbf{76.0} & 77.2 \\
    \bottomrule
  \end{tabular}}

  \label{fig:wins_ties}
\end{minipage}

\looseness=-1
\xhdr{State-of-the-art cofolding model distillation: \pearl~\citep{genesis2025pearl}}
We next explore applying our \nameshort framework to develop a few-step co-folding model by distilling the state-of-the-art model \pearl~\citep{genesis2025pearl}. We specifically adopt \pearldev, as a development checkpoint of \pearl trained with a \texttt{2023-12} cutoff date, and distill it into a student following the procedure in~\S\ref{sec:method}, yielding \nameshort-\pearl. In~\cref{tab:rnp_pearl} we compare \nameshort-\pearl to its teacher, to \nameshort-Boltz, and to several public frontier cofolding baselines on the \texttt{post-2023} subset of \runsnposes{} ($n{=}196$). \nameshort-\pearl matches the \pearldev teacher ($p=0.593$) on the key best@5 success rate while using $5{\times}$ fewer neural-network evaluations ($40$ NFEs vs. full simulation $\geq 200$ NFEs) and outperforms \pearldev on PB-valid ($p=0.034$). While the teacher retains a small edge on single-pose metrics (best@1 success rate and PB-validity), \nameshort-\pearl significantly outperforms all external baselines as well as \pearldev 10 diffusion steps and \nameshort-Boltz on these metrics across both best@1 and best@5. 
These results confirm that \nameshort{} can compress a diffusion-based cofolding model into a competitive student. 

\looseness=-1
The $5{\times}$ NFEs savings unlocked by \nameshort enable new workflows that remain impractical for full simulation. Concretely, for virtual screening, \nameshort-\pearl makes it feasible to cofold entire ligand libraries against a target of interest. In parallel, for synthetic data generation, the same speedup translates into an order-of-magnitude increase in the number of high-quality protein--ligand complexes that can be produced per unit of compute, which is a key bottleneck for training downstream scoring, affinity, and generative models. In both regimes, the success rate parity with the \pearldev teacher indicates that the diversity and quality of the top samples are preserved, and as a result, \nameshort-\pearl can be substituted for its teacher without sacrificing the structural signal that downstream applications depend on.

\subsection{Analysis of the compute-optimal frontier (Q2.)}
\label{sec:exp:compute_optimal}

\looseness=-1      
We next investigate the efficacy of \nameshort-Boltz in comparison to \boltz{1x} as a function of increasing inference budget on the \posebusters benchmark.  Through this study, we characterize the peak attainable performance 
of \boltz{1x} and \nameshort-Boltz. As such, we elucidate the precise inference recipe for \namealg that is optimal at each NFE budget.
We use \posebusters{} because its modest size (${\sim}300$ structures) makes the dense recipe-and-NFE sweep tractable. Moreover, its 2021-10-01 cutoff enables the principled investigation of the in-distribution Pareto frontier.

\begin{figure}[t]
    \centering
    \includegraphics[width=\linewidth]{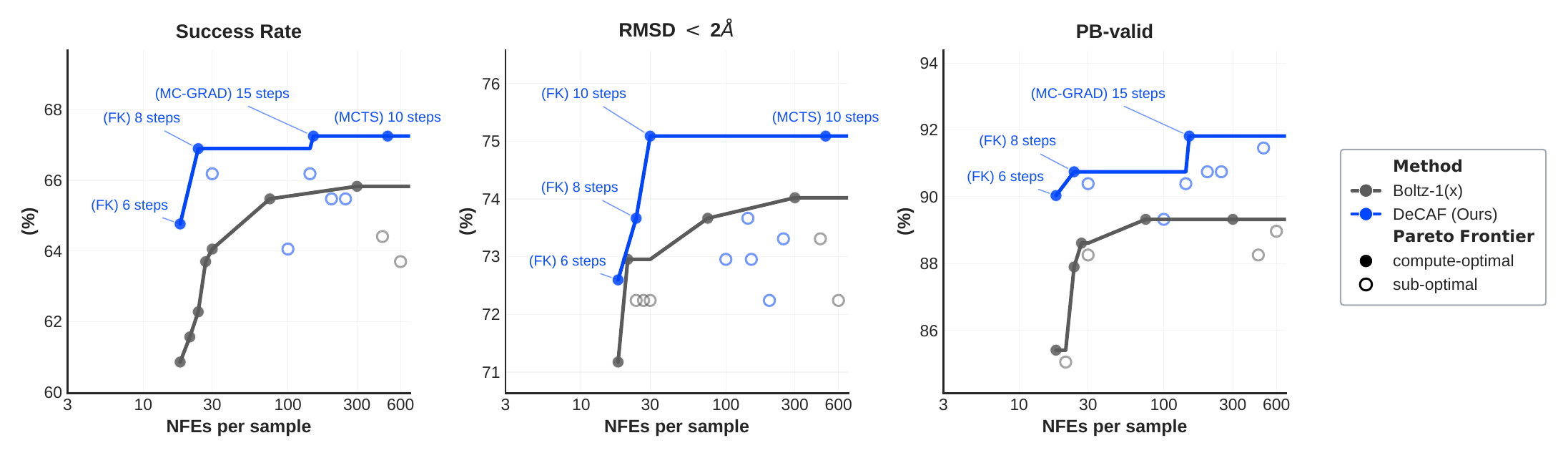}
    \vspace{-20pt}
    \caption{\looseness=-1 \small \textbf{\nameshort-Boltz{} is cost-effective as we increase compute.} A comparison of \namealg and \boltz{1x} as a function of NFE budget. The solid
    lines are the per-NFE compute optimal frontier for each method.}
    \vspace{-19pt}
    \label{fig:joint_nfe_scatter}
\end{figure}

\looseness=-1
\Cref{fig:joint_nfe_scatter} studies the frontier across PB Valid, RMSD$<\!2$\,\AA\, and Success Rate 
with a best $@5$ selection criterion for each NFE threshold. 
We find significant inference cost reductions for \namealg over \boltz{1x} full-scale configuration with 3 particles ($600$ NFEs) with as few as $30$ NFEs---\emph{a 20$\times$ inference cost reduction}. Importantly, owing to \nameshort-Boltz's flowmap lookahead, this reduction also leads to better quantitative performance with \namealg's Pareto frontier dominating \boltz{1x} across every NFE budget on all metrics. We further find that at different inference budgets, the exact Pareto-optimal recipe for \namealg varies. Specifically, at low NFEs ($\leq 30$), particle-based SMC akin to FK-steering is optimal. At the moderate NFEs ($100-250$), we find our Monte Carlo estimation of the reward gradient (MC-GRAD) to be the most effective. Finally, at large NFE budgets, we find \namealg with MCTS to be the most impactful at NFEs $\geq 142$.

\looseness=-1
\xhdr{Fine grained analysis}
As we increase the NFE based on the hyperparams in \namealg{}, however, we note that the head-to-head with \boltz{1x} favors different complexes, i.e., the successful complexes at high NFE are not a superset of the successful complexes at low NFE (\cref{tab:appendix_success_rates}). This suggests that a more complex interplay between sampling accuracy and reward guidance within \nameshort-Boltz. We highlight representative samples from \namealg (FK) at low NFE, which have higher accuracy than any pose sampled from \boltz{1x} at any NFE (\cref{fig:wins_ties}, additional samples in \cref{fig:method_grid}).

\looseness=-1
          
\subsection{Performance Analysis (Q3.)}
\label{sec:exp:qual}
      \begin{wraptable}{r}{0.45\linewidth}   %
      \vspace{-20pt}
        \centering
        \caption{\looseness=-1 \textbf{Parameterization ablation.}
        The $x_0$-aligned parameterization implements \cref{eq:denoiser_parameterization_flow_map}; the velocity parameterization predicts
        $v_\rho(x_\rho)$; consistency distillation (CD) follows \cite{song2023consistency}. Numbers on \posebusters, 30  NFEs.}
        \label{tab:x0_vs_velocity_pb202109}
        \small
        \setlength{\tabcolsep}{4pt}
        \begin{tabular}{lrrr}
        \toprule
        \textbf{Parametrization} &  \textbf{PB-Valid $\uparrow$} & \textbf{RMSD$<$2\AA\,$\uparrow$} \\
        \midrule
         $x_0$-aligned & \textbf{90.4} & \textbf{75.6}  \\
        velocity  & 9.2    & 49.1 \\
        Cons. Distillation & 0.0 & 49.5 \\
        \bottomrule
        \end{tabular}
        \vspace{-10pt}
    \end{wraptable}
    \looseness=-1
      \textbf{Posebusters validity.}
      We qualitatively study the pose quality at various NFE budgets. In~\cref{fig:wins_ties} we visually depict the samples and observe that, in comparison to \boltz{1x}, the MCTS version of \namealg{} ($600$ NFEs) can improve pose accuracy. Meanwhile, MC-GRAD ($150$ NFEs) reduces steric clashes in comparison to \boltz{1x}, highlighting improved physical reward alignment.

      \looseness=-1
      \namealg{} methods benefit from having both generally high pose quality (i.e., reward optimization) and more accurate ligand placement. While few-NFE inference with \nameshort-Boltz already yields better failure rates in relation to \boltz{1x} (c.f.~\cref{tab:appendix_subcheck}) we see further refinement in pose quality when scaling NFEs further with the MC-GRAD and MCTS variants. We additionally note that pose quality on \posebusters is greatly determined by reward design, and leads to common failure modes across \nameshort-Boltz and \boltz{1x} that share the same potentials. For instance, we observe that sp2-hybridized bonds are often not planar. In addition, some \posebusters quality checks have stricter tolerances than are chemically accurate, such as flagging metal ion coordination as a "clash" based solely on the neutral atoms' van der Waals radii (\cref{fig:method_grid_pb_fail}). We find majority of these failures can be mitigated through inference search.

    \looseness=-1
      \xhdr{Generalization  on chemically-relevant targets}
        We stratify the \runsnposes benchmark to evaluate our models on challenging scenarios most relevant to drug discovery tasks. We consider several slices which probe these settings: drug-like ligands only, ligands interacting with ions and cofactors, and ligands at protein-protein interfaces. In these subsets, \namealg outperforms \boltz{1x} to a statistically significant degree (\cref{tab:appendix_chem_rel}). We curate representative poses in~\cref{fig:method_grid_druglike}.

    \looseness=-1
    \xhdr{Parametrization}
We ablate \nameshort{}'s denoiser-aligned parametrization
(\cref{eq:denoiser_parameterization_flow_map}) against a velocity-predictor student~\citep{geng2025improved} that directly regresses $v_{\rho}(x_\rho)$ under the same JVP schedule and data budget. We also experimented with consistency distillation parametrization~\citep{song2023consistency}. ~\Cref{tab:x0_vs_velocity_pb202109} confirms that only our chosen $x_0$-aligned parametrization can sample valid poses. The denoiser parametrization reuses EDM preconditioning and enables Kabsch alignment on $\hat{x}_0$---which we find critical for stable training.

\section{Related Work}
\label{sec:related_work}

\begin{figure}[t]
    \centering
    \includegraphics[width=\linewidth]{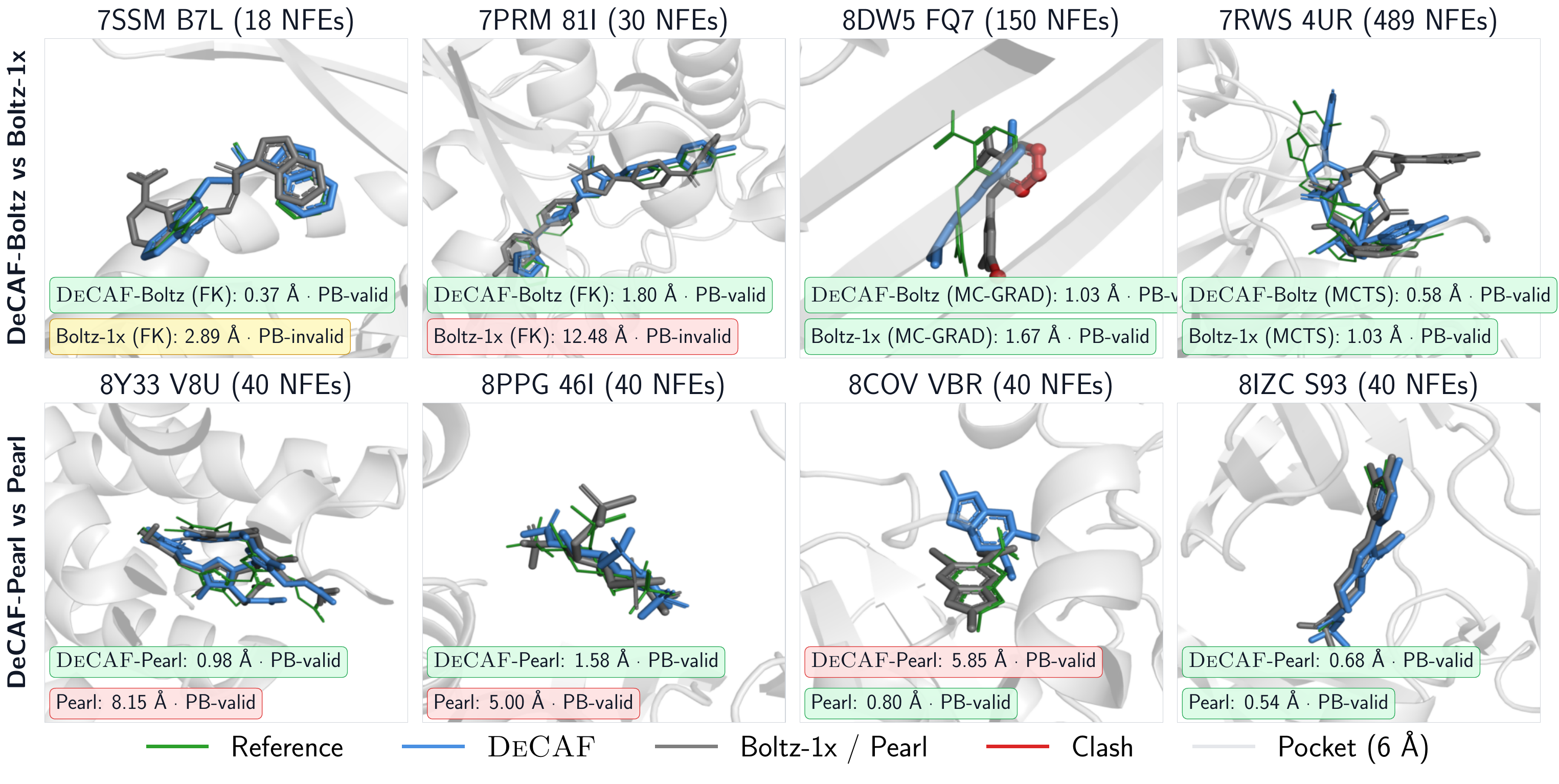}
    \vspace{-20pt}
    \caption{\small \looseness=-1 
    Each panel overlays the \textcolor[RGB]{44, 160, 44}{\textbf{ground-truth}} crystal ligand against \textcolor[RGB]{74, 144, 226}{\textbf{\namealg}} and \textcolor[RGB]{127, 127, 127}{\textbf{\boltz{1x}}} (first row) and \textcolor[RGB]{127, 127, 127}{\textbf{\pearl}} (second row) samples at the specified NFE. Row~1 compares \namealg{}-Boltz against \boltz{1x} and Row~2 compares \namealg{}-\pearl against \pearl. The protein pocket is shown as a light-gray cartoon, and predicted ligand atoms that \textcolor[RGB]{220, 38, 38}{clash} with the protein in red. At the bottom of each panel, we report the RMSD and posebusters' validity of each pose.}
    \vspace{-17pt}
    \label{fig:wins_ties}
\end{figure}

\looseness=-1
\xhdr{Protein generation}
Early generative models focused on backbone design~\citep{yim2023fast,watson2023novo,bose2023se,huguet2024sequence,geffner2025proteina,geffner2025laproteina}.
Since \alphafold{3}~\citep{abramson2024accurate} established EDM-style denoising~\citep{karras2022elucidating} over all-atom coordinates as the dominant paradigm for cofolding, several open and closed-source systems have followed, including \boltz{1(x)}~\citep{wohlwend2025boltz}, \protenix~\citep{protenixv1}, \chai{1}~\citep{chai2024chai1}, \pearl~\citep{genesis2025pearl}, Complexa~\citep{didi2026scaling}, and DISCO~\citep{rector2026general}. These all share a costly inference-time bottleneck of $\mathcal{O}(200)$ NFEs per sample. %

\looseness=-1
\xhdr{Few-step generative models}
Several methods compress full generative trajectories, including consistency models~\citep{song2023consistency,song2023improved}, Consistency Trajectory Models~\citep{consistencyTrajectory}, shortcut models~\citep{shortcut}, and the MeanFlow~\citep{meanflows,boffi2024flow}. Despite this flourishing literature, the distillation of all-atom cofolding models remains underexplored, with only DCFold as closed-source concurrent work \citep{zhang2026dcfold}.

\looseness=-1
\xhdr{Inference-time steering}
Inference-time techniques for diffusion include classifier guidance~\citep{ho2022classifier}, Universal Guidance~\citep{bansal2023universalguidancediffusionmodels}, and particle-based methods derived from SMC~\citep{Del-Moral:2006}, including Feynman--Kac methods~\citep{singhal2025general,skreta2025feynman} and diffusion MCTS~\citep{jain2025diffusion}. In the protein setting, \af{3}-family models routinely employ physics-informed potentials such as the \boltz{1x} stereochemical potentials~\citep{wohlwend2025boltz} whose cost grows linearly in both denoising steps and particle count. %

\section{Conclusion}
\label{sec:conclusion}
\looseness=-1
We introduced \nameshort, a flow map that distills a pretrained all-atom cofolding diffusion model into a few-step generator. The construction rests on two technical choices: a reparameterization in $\sigma$-space that matches the EDM-style adopted by standard cofolding models, and a denoiser parametrization that preserves the $\sethree$ rigid alignment. On top of \nameshort, we built \namealg, an inference-time search framework that leverages flow map lookahead to achieve higher-fidelity reward alignment. Empirically, \namealg matches the $600$-NFE \boltz{1} teacher with a $20\times$ reduction in function evaluations on \posebusters and improves over \boltz{1x} on \runsnposes at every low-NFE setting we considered. Furthermore, we distilled the state-of-the-art cofolding model \pearldev into \nameshort-\pearl. We found that \nameshort-\pearl achieved state-of-the-art performance, outperforming diffusion-based models and baselines, and matching the success rate of its teacher with $5\times$ fewer diffusion NFEs.

\looseness=-1
\xhdr{Limitations} Several limitations point to the next steps. Our reward signal is inherited from \boltz{1x}, so failure modes such as non-planar sp\textsuperscript{2} bonds reflect this choice rather than the search procedure. The $\sigma$-space denoiser formulation is not specific to cofolding and may extend to nucleic acids, larger assemblies, and multi-chain systems. Finally, the non-monotone relationship between NFE budget and per-target success suggests that adaptive search and joint training of the flow map with task-specific rewards are natural avenues for further gains. 

\section*{Acknowledgments}
\looseness=-1
The authors would like to thank Matthew Wicker, Zhengrui Xiang, and Ken Leidal for helpful feedback on early drafts of this work. In addition, we would like to thank Hannes St$\ddot{\text{a}}$rk for helpful advice about the Boltz code base.
J.N. acknowledges support from the Mathworks Fellowship. R.S. and P.H. acknowledge support from the Machine Learning for Pharmaceutical Discovery and Synthesis (MLPDS) consortium, DSO Singapore grant on next generation techniques for protein ligand binding, and a grant from Siemens Corp on inverse design. GS, PM, and MA would also like to thank the entire Genesis Research team for the valuable support. 

\clearpage

\bibliography{bibliography}
\bibliographystyle{abbrvnat}
\newpage
\appendix

\section*{Appendix}

\section{\namealg}
\label{app:sampling_algos}

\begin{algorithm}[H]

\caption{Inference-time search with \namealg}
\label{alg:fm_search}
\begin{algorithmic}[1]
\Require FlowMap $X$; noise schedule $\sigma_N > \sigma_{N-1} > \cdots > \sigma_0 = 0$;
  particles $P$; resampling interval $L$; steering weight $\lambda$; number of search iterations $S$;
  potential $R$;
  $\gamma \in [0,1]$; 
  rollout horizon $K$ (with $K=1$ for FK).
\Ensure Sample $\hat{x}_0$.
\State Draw $\{x_{\sigma_N}^{(p)}\}_{p=1}^{P} \sim \mathcal{N}(0, \sigma_N^2 I)$ \Comment{initialize $P$ particles}
\State $\mathcal{X} \gets \{x_{\sigma_N}^{(p)}\}_{p=1}^{P}$,\quad $n \gets 0$ \Comment{global search tree, current noise index}
\For{$s = 0, 1, \ldots, S-1$}
    \State $\mathcal{T} \gets \emptyset$ \Comment{rollout trajectories}
    \State $n_{\text{end}} \gets \min(N,\; n + K)$
    \For{$m = n, \ldots, n_{\text{end}} - 1$} \Comment{denoise rollout (all particles in parallel)}
        \State $\hat{x}_0^{(p)} \gets X(x_{\sigma_m}^{(p)}, \sigma_m, 0)$, \quad $\forall p$
        \State \textit{Optional:} gradient steps on $\hat{x}_0^{(p)}$ w.r.t.\ $R$ \Comment{see \cref{eq:gradient_steps}}
        \State $\tilde{\sigma}_{m+1} \gets \sqrt{1-\gamma^2}\,\sigma_{m+1}$ \Comment{$\gamma$ renoise}
        \State $\tilde{x}^{(p)} \gets \hat{x}_0^{(p)} + \tilde{\sigma}_{m+1}\,\dfrac{x_{\sigma_m}^{(p)} - \hat{x}_0^{(p)}}{\sigma_m}$, \quad $\forall p$
        \State $\epsilon^{(p)} \sim \mathcal{N}(0, I)$, \quad $\forall p$
        \State $x_{\sigma_{m+1}}^{(p)} \gets \tilde{x}^{(p)} + \gamma\,\sigma_{m+1}\,\epsilon^{(p)}$, \quad $\forall p$
        \State $\mathcal{T} \gets \mathcal{T} \cup \bigl\{(x_{\sigma_{m+1}}^{(p)},\, m+1)\bigr\}_{p=1}^{P}$
    \EndFor
   \State Compute particle scores $R^{(p)}\bigl(\hat{x}_0^{(p)}\bigr)$, $\forall p$
    \If{$\textsc{search} = \textsc{MCTS}$}
        \State $\mathcal{X} \gets \mathcal{X} \cup \mathcal{T}$
        \State Update weights of all $(x, m) \in \mathcal{T}$ \Comment{backup}
        \State Draw $(x_{\sigma_i}^{(p)}, i) \sim \mathcal{X}$ according to UCT, $\forall p$
    \ElsIf{$\textsc{search} = \textsc{FK}$} \Comment{$K=1$, so $\mathcal{T}$ holds one entry per particle}
        \State $(x_{\sigma_i}^{(p)}, i) \gets$ the entry of $\mathcal{T}$ for particle $p$, $\forall p$
        \If{$s \bmod L = 0$}
            \State Resample $\{x_{\sigma_i}^{(p)}\}_{p=1}^{P}$ with weights $\propto R^{(p)}$
        \EndIf
    \EndIf
    \State $n \gets i + 1$ \Comment{advance to next noise level}
\EndFor
\State \Return $x_0^{(p^\star)}$ with $p^\star = \arg\max_p \bigl\{ R(x_0^{(p)}) : (x_0^{(p)}, 0) \in \mathcal{X} \bigr\}$

\end{algorithmic}
\end{algorithm}

\looseness=-1
\xhdr{\namealg (MC-Grad)} As a slight extension of \cref{alg:fm_search}, we realized that the gradient estimate in $\nabla_{x_{\sigma_0}} R(x_{\sigma_0})$ can be a noisy estimate of the optimal guidance direction. We hypothesize that a Monte Carlo average of several gradients might be more favorable, as observed previously \citep{holderrieth2026diamond}. In order to do so, we set
\begin{align}
x_{0}\leftarrow x_{0} + \frac{\beta}{L}\sum\limits_{l=1}^{L}w_l\nabla_{x_{0}^l} R(x_{0}^l)
\end{align}
where weights $w_{l}$ are the softmax of importance logits derived from the local reward and the renoise prior, $\beta =(\sigma/\sigma_\text{data}^{2})$, and $x_{0}^l$ is obtained by renoising $x_{\sigma}$ back to $x_{\tilde\sigma}$ and then performing the rollouts step in line $7$ (i.e. this induces one extra loop that we omit here for readability).

\section{Experimental Setup}
\label{app:experimental_setup}

    \begin{table}
     \centering
    \caption{Shared inference-parameter settings (constant across NFE budgets and methods).}
    \label{tab:hyper_shared}
    \small
    \begin{tabular}{@{}ll@{}}
      \toprule
      Parameter & Value \\
      \midrule
      EDM noise schedule & Karras et al.\ ($\rho{=}7$) \\
      \quad $\sigma_\mathrm{min}$, $\sigma_\mathrm{max}$, $\sigma_\mathrm{data}$ & $0.0004$, $160$, $16$
   \\
      Stochastic-step noise scale (\texttt{noise\_scale}) & $0.901$ \\
      \midrule
    Per-step gradient updates            & $20$ \\   
      \midrule
      MCTS selection rule                       & UCT \\
      UCT exploration constant ($c$) & $1.0$ \\
      Progressive widening ($k$, $\alpha$) & $2.0$, $0.5$ \\
               
      \bottomrule
    \end{tabular}
  \end{table}

  \begin{table}[h]
    \centering
    \caption{Per-method settings on \runsnposes{} and \posebusters{}.
    \texttt{Steps}: number of sampling steps. \texttt{Params}: per-method
    hyperparameters that scale the per-pose NFE.
    All other sampler hyperparameters are fixed across NFE budgets.}
    \label{tab:hyper_nfe}
    \small
    \setlength{\tabcolsep}{6pt}
    \begin{tabular}{@{}lccc@{}}
      \toprule
      Method & Steps & Params & NFE per pose \\
      \midrule
      \multicolumn{4}{@{}l}{\textit{\runsnposes{} (4 FK particles)}} \\
      \boltz{1x}\,/\,\nameshort{} (FK)        & $2$   & $4$ particles                       & $8$ \\
      \boltz{1x}\,/\,\nameshort{} (FK)        & $5$   & $4$ particles                       & $20$ \\
      \boltz{1x}\,/\,\nameshort{} (FK)        & $6$   & $4$ particles                       & $24$ \\
      \boltz{1x}\,/\,\nameshort{} (FK)        & $10$  & $4$ particles                       & $40$ \\
      \boltz{1x}\,/\,\nameshort{} (FK)        & $12$  & $4$ particles                       & $48$ \\
      \namealg(MC-GRAD)                     & $15$  & $10$ samples      & $150$ \\
           \boltz{1x}        & $40$  & $4$ particles                       & $160$ \\
      \boltz{1x} \emph{(ref.)}                & $200$ & $4$ particles                       & $800$ \\
      \namealg(MCTS)                     & $10$  & $4$ children, $50$ simulations      & $800$ \\
      \midrule
      \multicolumn{4}{@{}l}{\textit{\posebusters{} (3 FK particles)}} \\
      \boltz{1x}                         & $[6, \dots, 200]$   & $3$ particles                       & $[18, \dots, 600]$ \\

      \boltz{1x} \emph{(ref.)}           & $200$ & $3$ particles                       & $600$ \\
      \addlinespace[2pt]
      \nameshort{} (FK)                       & $6$   & $3$ particles                       & $18$ \\
      \nameshort{} (FK)                       & $8$   & $3$ particles                       & $24$ \\
      \nameshort{} (FK)                       & $10$  & $3$ particles                       & $30$ \\
      \nameshort{} (MC-GRAD)                     & $10$  & $10$ samples                        & $100$ \\
            \nameshort{} (MCTS)                     & $10$  & $5$ sims, $4$ children              & $142$ \\
      \nameshort{} (MC-GRAD)                     & $15$  & $10$ samples                        & $150$ \\
      \nameshort{} (MC-GRAD)                     & $20$  & $10$ samples                        & $200$ \\
      \nameshort{} (MC-GRAD)                     & $25$  & $10$ samples                        & $250$ \\
      \nameshort{} (MCTS)                     & $10$  & $15$ sims, $4$ children             & $489$ \\
      \bottomrule
    \end{tabular}
  \end{table}

\looseness=-1

\looseness=-1

\subsection{Architecture and Training}
  We adopt Boltz-1 original architecture and codebase \citep{wohlwend2025boltz} except for the implementation of dual-time conditioning. The denoiser parametrization $u(x_\rho, \rho, \sigma)$ requires fusing two noise levels into the score network's conditioning stream. We adopt a dual-time conditioning module from FastGen \citep{fastgen2026} that departs from the single-time conditioning of the EDM teacher. We train \nameshort as a distillation head with our $x0$-aligned loss (\cref{eq:main_loss}), while the trunk and EDM modules are frozen and initialized from the \boltz{1} open-source checkpoint. Training uses RCSB PDB structures released before 2021-09-30, filtered to resolution $\le$ 9.0 Å. We train for 100 epochs with 51,200 samples per epoch on 64 H200 GPUs, using a per-GPU batch size of 2 (effective batch size 128) and diffusion multiplicity 16. The optimizer is Muon with momentum 0.95 and Nesterov updates, with an AdamW fallback ($\beta_1=0.9$, $\beta_2=0.95$) for 1D parameters at learning rate ratio 0.1; weight decay is 0.01 throughout. The learning rate follows an
  \alphafold{3}-style schedule with linear warmup over 1,000 steps to a peak of $1.8 e^{-3}$. The $\sigma$-weighting follows the Karras EDM schedule ($\sigma_\text{min}=4e^{-4}$, $\sigma_\text{max}=160$, $\rho=7$, $\sigma_\text{data}=16$), with 10\% diagonal samples ($\sigma_r=\sigma_t$) and 90\% off-diagonal. We further re-scale loss $\mathcal{L}$ in \cref{eq:main_loss} by $\frac{1}{\sqrt{(\mathcal{L} + 1e^{-6})}}$

\subsection{Hyperparameters}

We report the inference hyperparameters in tables \ref{tab:hyper_shared} and \ref{tab:hyper_nfe}. Values in \cref{tab:hyper_shared} are
  constant across all NFE budgets and across both methods unless otherwise
  noted. Values in \cref{tab:hyper_nfe}
  are the only knobs that change with the compute budget.

The key sampler tuning we apply to \boltz{1x} in the few-step regime is
  setting the step scale $\sigma_\mathrm{scale} = 1.0$ (vs.\ the default
  $1.638$ used at $200$ steps). With the default $1.638$, the SDE diverges
  under aggressive step-size schedules, producing $0.0\%$ on every metric
  at $2$ steps (\cref{tab:rnp}, ``\boltz{1x} (default)''). Setting noise scale
  $\eta=1.0$ recovers stable sampling and is what we
  report as ``\boltz{1x} (tuned)'' throughout.

\section{Additional Ablations}
\label{app:additional_experiments}
  \subsection{Sampler ablations}
  \label{sec:sampler-ablations}

  \begin{figure}[t]
      \centering
      \begin{minipage}[h]{0.48\textwidth}
          \centering
          \captionof{table}{Sweep of $\gamma$ for Alg.~\ref{alg:fm_search} on \posebusters{}. 10 sampling steps, \namealg (FK), $P=3$.}
          \label{tab:ctm-gamma-sweep}
          \small
          \setlength{\tabcolsep}{4pt}
          \begin{tabular}{lrrr}
              \toprule
              $\gamma$ & RMSD$<\!2$ $\uparrow$ & PB Valid $\uparrow$ & Success Rate $\uparrow$\\
              \midrule
              $0.3$ & 75.3 & 89.2 & 65.2 \\
              $0.5$ & \textbf{75.6} & 90.3 & \textbf{65.9} \\
              $0.7$ & 72.8 & 90.3 & 63.8 \\
              $1.0$ & 70.3 & \textbf{90.7} & 61.3 \\
              \bottomrule
          \end{tabular}
      \end{minipage}
      \hfill
      \begin{minipage}[h]{0.48\textwidth}
          \centering
          \includegraphics[width=.6\linewidth]{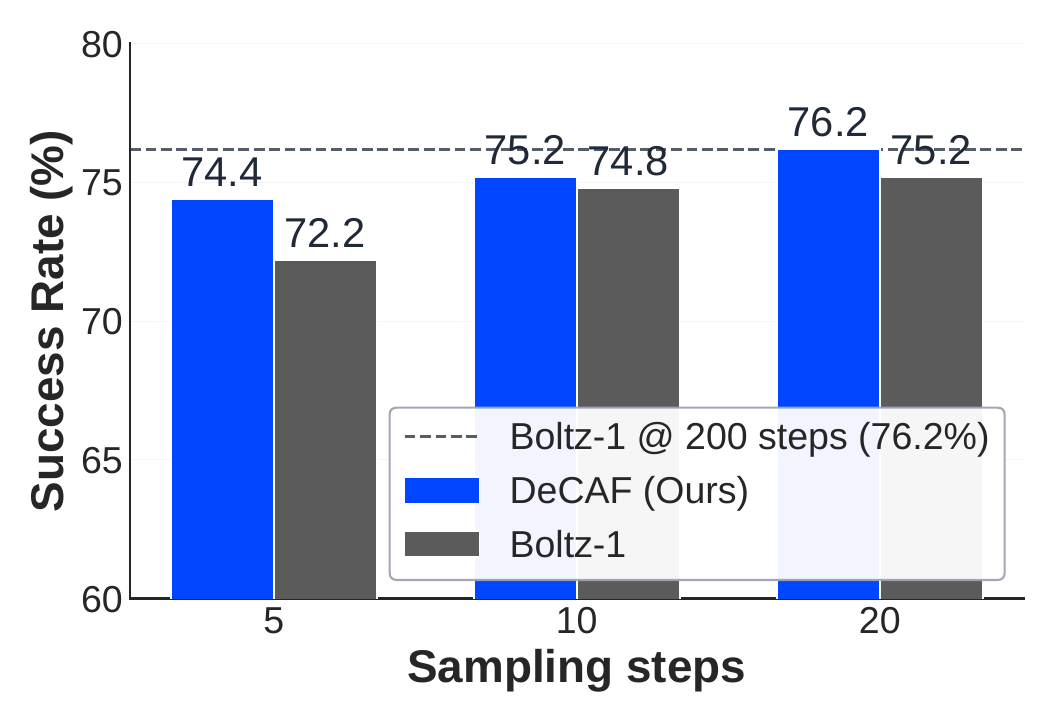}
          \captionof{figure}{\small \looseness=-1 RMSD$<\!2$\,\AA{} best@5 on \posebusters. \nameshort{} (FlowMap sampler Alg. \ref{alg:gamma_sampler}, $\gamma{=}0$) vs \boltz{1} (ODE) at matched sampling steps; dashed line marks \boltz{1} at 200 steps.}
          \label{fig:rmsd2-step-hist}
      \end{minipage}
  \end{figure}

  \paragraph{Stochasticity $\gamma$ sweep.}
\Cref{tab:ctm-gamma-sweep} sweeps the CTM stochasticity parameter $\gamma \in \{0.3, 0.5, 0.7, 1.0\}$ for \cref{alg:fm_search} on \posebusters{}, holding the rest of the sampler fixed (10 denoising steps, \namealg (FK), $P=3$ FK particles). We adopt $\gamma=0.5$, which we hypothesize balances the determinism of $\gamma\!\to\!0$ against the variance of sampling at $\gamma{=}1$. Empirically, Success Rate peaks at $\gamma{=}0.5$ (65.9\%), with a 4.7-pt spread between the best ($\gamma{=}0.5$) and worst ($\gamma{=}1.0$) settings, while PB-validity is essentially saturated across all $\gamma$ ($\Delta\!\leq\!1.5$\,pp).

  \paragraph{\nameshort{} vs \boltz{1} without inference scaling.}
  \Cref{fig:rmsd2-step-hist} isolates the sampler axis: both methods run unguided ODE-style integration so the comparison reflects the underlying sampler quality in isolation rather than any
  inference-time augmentation. \nameshort{}'s FlowMap sampler (taking velocity predictions directly from the trained flow map at each step) outperforms \boltz{1}'s standard ODE solver at every sampling-step count — at 5 / 10 / 20 steps respectively by
  $+2.2$, $+0.4$, and $+1.0$\,pp on RMSD$<\!2$\,\AA{}. Most notably, \nameshort{} at 20 steps (76.2\%) \emph{matches} \boltz{1} at 200 steps (76.2\%, dashed reference), a $10\times$ reduction in inference compute.

\begin{table}[t]
  \centering
  \caption{\textbf{Success rate breakdown for \namealg as NFEs grow.} We report the fraction of \runsnposes{} that succeed along the \namealg Pareto front, highlighting that while increasing NFEs does improve accuracy, the low-NFE variants do offer complementary successes to the most expensive MCTS-based sampling method.
  }
  \label{tab:appendix_success_rates}
  \small
  \setlength{\tabcolsep}{4pt}
  \begin{tabular}{lrr}
    \toprule
    Outcome & count & \% \\
    \midrule
    All three succeed & 479 & 68.23\% \\
    Only \nameshort FK (NFE=40) + \nameshort MCTS (NFE=800) succeed & 24 & 3.42\% \\
    Only \nameshort FK (NFE=20) + \nameshort MCTS (NFE=800) succeed & 7 & 1.00\% \\
    Only \nameshort FK (NFE=20) + \nameshort FK (NFE=40) succeed & 7 & 1.00\% \\
    Only \nameshort MCTS (NFE=800) succeeds & 25 & 3.56\% \\
    Only \nameshort FK (NFE=40) succeeds & 7 & 1.00\% \\
    Only \nameshort FK (NFE=20) succeeds & 7 & 1.00\% \\
    None succeeds & 146 & 20.80\% \\
    \midrule
    Total & 702 & 100.00\% \\
    \bottomrule
    \end{tabular}
\end{table}
\begin{table}[t]
  \centering
  \caption{\textbf{Per-complex failure rates by \posebusters{} sub-check}
  (best@5 over \posebusters metrics). \namealg exceeds \boltz{1x} at 600 NFE across all NFE levels, and we see generally increasing pose quality with increasing NFE. Lower is better for all entries in this table.}
  \label{tab:appendix_subcheck}
  \small
  \setlength{\tabcolsep}{4pt}
  \resizebox{\textwidth}{!}{%
  \begin{tabular}{lccccc}
    \toprule
    \textbf{Sub-check} & \multicolumn{4}{c}{\textbf{\namealg}} & \multicolumn{1}{c}{\textbf{\boltz{1x}}}\\
    & NFE = 20 & NFE = 40 & NFE = 75 & NFE = 489 & NFE = 600 \\
\midrule
double bond flatness & 6.34\% & 8.10\% & 7.27\% & 7.29\%  & 7.83\% \\
volume overlap with protein & 2.82\% & 2.11\% & 0.69\% & 2.43\% &  3.20\% \\
bond angles & 3.87\% & 2.11\% & 0.35\% & 0.00\% &  0.00\% \\
bond lengths & 1.76\% & 1.41\% & 0.00\% & 0.00\% &  0.00\% \\
volume overlap with organic cofactors & 1.06\% & 1.76\% & 0.00\% & 1.07\% & 1.07\% \\
minimum distance to protein & 0.35\% & 0.00\% & 0.35\% & 0.00\% &  0.36\% \\
minimum distance to organic cofactors & 0.00\% & 0.00\% & 0.35\% & 0.00\% &  0.36\% \\
minimum distance to waters & 0.00\% & 0.00\% & 0.35\% & 0.00\% &  0.36\% \\
internal steric clash & 0.70\% & 0.00\% & 0.00\% & 0.00\% &  0.00\% \\
\midrule
\textbf{PB-valid pass rate} & 89.79\% & 90.14\% & \textbf{92.04\%} & \underline{90.62\%} & 88.97\% \\
\bottomrule
\end{tabular}}
\end{table}

\begin{table}[t]
  \centering
  \caption{\textbf{Method performance on challenging and chemically-relevant targets.} We measure best@5 joint \rmsdtwoa and PB-valid performance on the following slices of the \runsnposes set (higher is better). Drug-likeness is determined using a Quantitative Estimate of Drug-likeness (QED) score threshold of 0.65, as is typical for a drug-like small molecule \citep{Bickerton2012}. Methods marked with an asterisk have a statistically-significant improvement relative to the best \boltz{1x} version as measured by a two-sided paired Wilcoxon signed-rank test with $p < 0.01$.
  }
  \label{tab:appendix_chem_rel}
  \small
  \setlength{\tabcolsep}{4pt}
  \begin{tabular}{@{}lrrrrr@{}}
    \toprule
    \textbf{Slice} & \textbf{N} & \multicolumn{2}{c}{\textbf{\namealg}} & \multicolumn{2}{c}{\textbf{\boltz{1x}}} \\
    && NFE = 40 & NFE = 800 & NFE = 40 & NFE = 800\\
    \midrule
    Drug-like ligands & 283 & **\textbf{0.809} & 0.806 & 0.787  & 0.783 \\
    Ion/cofactor coordination & 134 & 0.754  & **\textbf{0.761} & 0.749 & 0.760 \\
    Ligand at PPI & 292 & 0.777 & **\textbf{0.784} & 0.779 & 0.780 \\
    \bottomrule
  \end{tabular}
\end{table}

\subsection{Qualitative analysis}
\label{app:method_grid}

\begin{figure}[p]
  \centering
  \includegraphics[width=\linewidth]{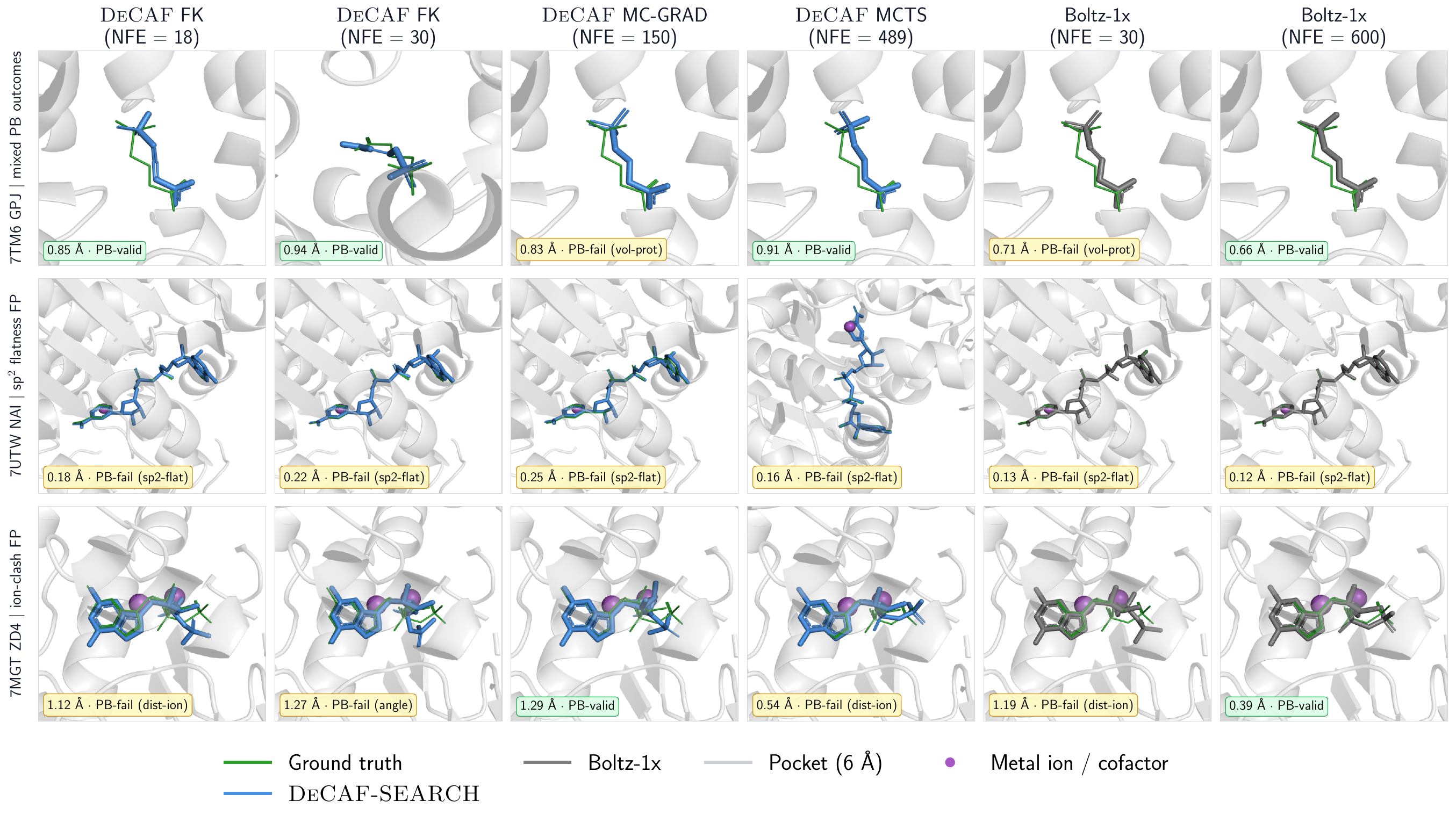}
  \caption{\textbf{Example failure modes due to \posebusters checks}
\nameshort{} predictions blue, Boltz-1 gray. Pocket cartoon (gray) is
  the residues within 6\,\AA{} of the crystal ligand (green).
  }
  \label{fig:method_grid_pb_fail}
\end{figure}

\paragraph{\posebusters{} quality checks.}
\Cref{tab:appendix_subcheck} collates statistics of the different \posebusters{} sub-check failures for \namealg and \boltz{1x}. As noted in the main text \cref{sec:exp:qual}, we see high failure rates due to lack of sp2-hybridized bond flatness across all models, likely due to suboptimal reward design. Other checks are failed much less frequently and generally improve with increasing NFE (consider e.g. the bond angles failure mode). In \cref{fig:method_grid_pb_fail} we highlight some notable \posebusters failures, including examples of the pervasive sp2 planarity issue (middle row). We also flag some false positives of the \posebusters checks, where accurate poses which capture ionic coordination are flagged as having incorrect distance and volume overlap.

\begin{figure}[p]
  \centering
  \includegraphics[width=\linewidth]{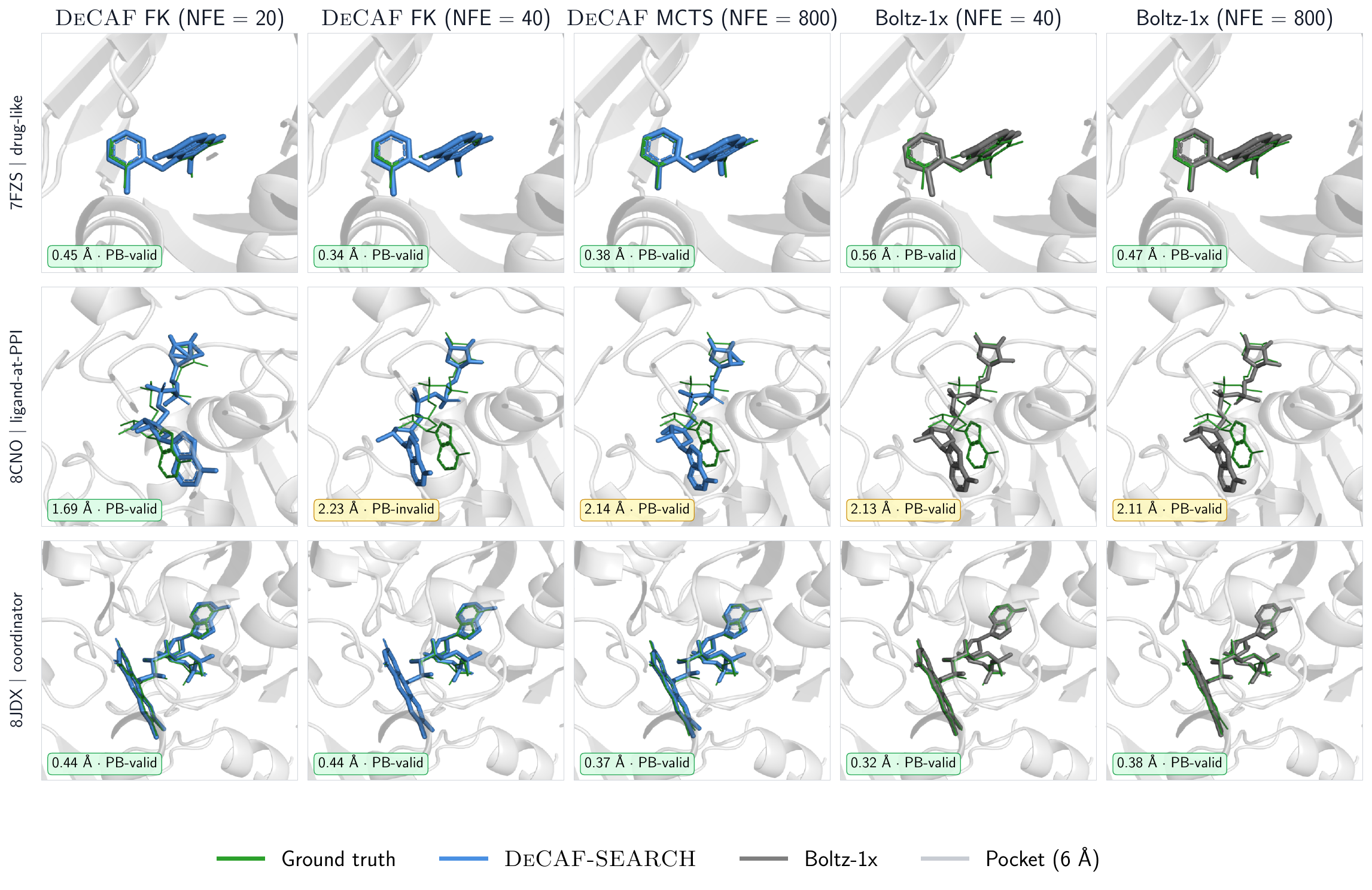}
  \caption{\textbf{Qualitative grid for chemically-relevant subsets of \runsnposes{}.}
\nameshort{} predictions blue, Boltz-1 gray. Pocket cartoon (gray) is
  the residues within 6\,\AA{} of the crystal ligand (green).
  }
  \label{fig:method_grid_druglike}
\end{figure}

\paragraph{Practically-relevant slices of \runsnposes.}
As we note in \cref{sec:exp:qual}, it is critical to validate the performance of \nameshort on systems that are relevant for users of cofolding models. To emulate this setting, we slice \runsnposes{} to a subset of structures that highlight common settings in small-molecule drug discovery and report results in \cref{tab:appendix_chem_rel}. We note that \namealg has strong performance at both low and high NFE, and matches or outperforms \boltz{1x} at similar NFEs. We also highlight select structures from each category in \cref{fig:method_grid_druglike}.

\begin{figure}[p]
  \centering
  \includegraphics[width=\linewidth]{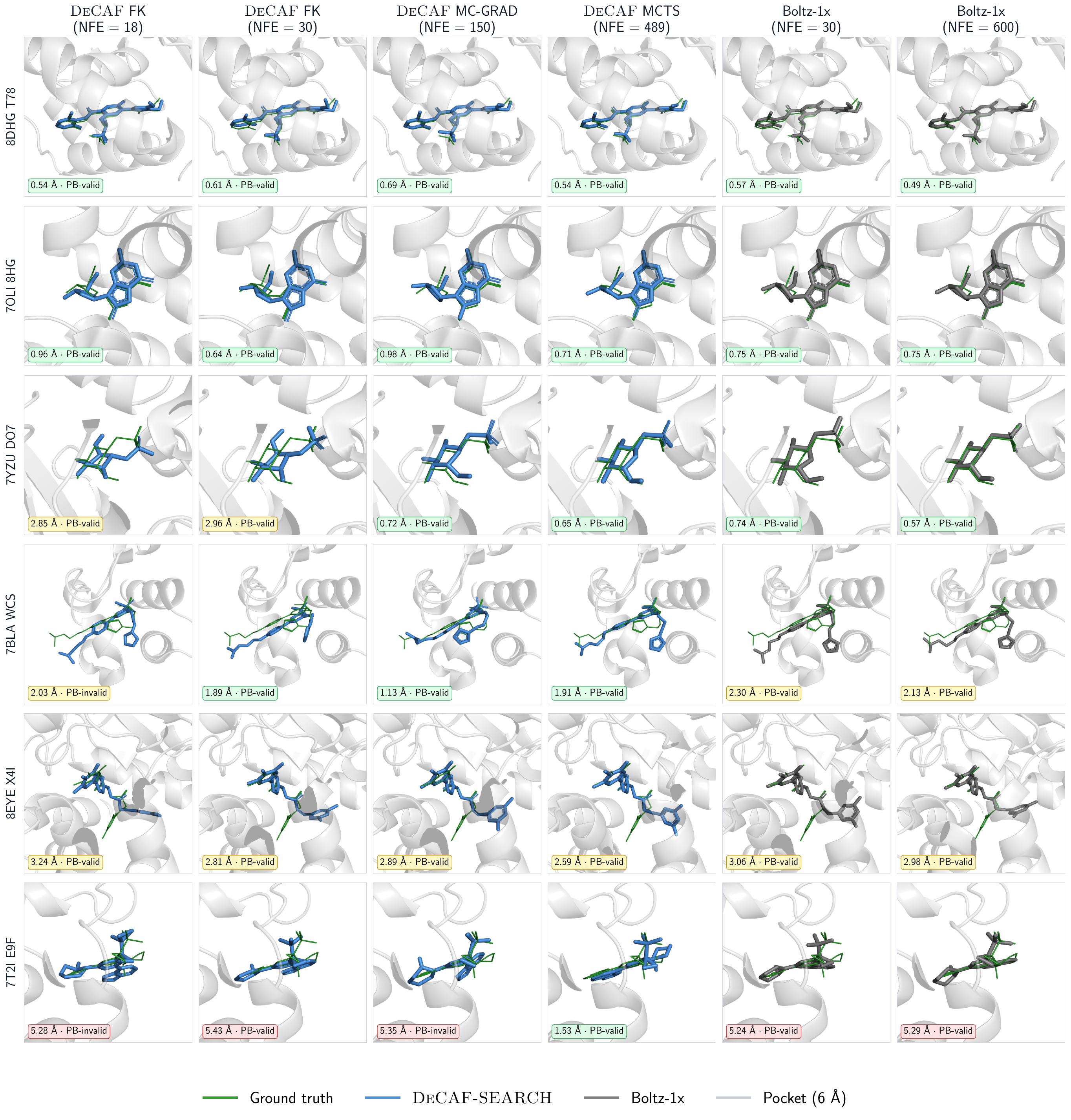}
  \caption{\textbf{Multi-method qualitative grid for \posebusters{} complexes.}
\nameshort{} predictions blue, Boltz-1 gray. Pocket cartoon (gray) is
  the residues within 6\,\AA{} of the crystal ligand (green).
  }
  \label{fig:method_grid}
\end{figure}

\newpage

\clearpage

\end{document}